\documentclass[twoside,11pt]{article}
\usepackage{amsfonts}
\usepackage{mathrsfs}
\usepackage[utf8]{inputenc} 
\usepackage{float}
\usepackage{enumitem}
\usepackage{amsmath,amssymb}

\usepackage{latexsym}
\usepackage{longtable}
\usepackage{makecell}
\usepackage{mathrsfs}

\usepackage{booktabs}
\usepackage{epstopdf}
\usepackage{tabularx}
\usepackage{tikz}
\usetikzlibrary{shapes.geometric, arrows}
\usetikzlibrary{positioning}
\tikzstyle{startstop} = [rectangle, rounded corners, minimum width=3cm, minimum height=1cm, text centered, draw=black, fill=red!30]
\tikzstyle{process} = [rectangle, minimum width=3cm, minimum height=1cm, text centered, draw=black, fill=blue!30]
\tikzstyle{io} = [trapezium, trapezium left angle=70, trapezium right angle=110, minimum width=3cm, minimum height=1cm, text centered, draw=black, fill=green!30]
\tikzstyle{arrow} = [thick,->,>=stealth]
\usepackage{geometry}  
\usepackage[active]{srcltx}
\usepackage{graphicx}
\usepackage{booktabs}
\usepackage{multirow}
\usepackage{siunitx}
\usepackage{xurl} 
\usepackage{amsmath}
\usepackage{setspace}
\usepackage{graphicx} 
\usepackage{algorithm}
\usepackage{algpseudocode}

\usepackage{subcaption} 
\usepackage[
    bookmarks=true,         
    bookmarksnumbered=true, 
    colorlinks=true, pdfstartview=FitV, linkcolor=blue, citecolor=blue,
    urlcolor=blue]{hyperref}

\usepackage{fancyhdr}
\usepackage{lipsum} 
\usepackage{listings} 
\usepackage{xcolor} 
\definecolor{codegreen}{rgb}{0,0.6,0}
\definecolor{codegray}{rgb}{0.5,0.5,0.5}
\definecolor{codepurple}{rgb}{0.58,0,0.82}
\definecolor{backcolour}{rgb}{0.95,0.95,0.92}

\lstdefinestyle{mystyle}{
    backgroundcolor=\color{backcolour},
    commentstyle=\color{codegreen},
    keywordstyle=\color{magenta},
    numberstyle=\tiny\color{codegray},
    stringstyle=\color{codepurple},
    basicstyle=\footnotesize\ttfamily,
    breakatwhitespace=false,
    breaklines=true,
    captionpos=b,
    keepspaces=true,
    numbers=left,
    numbersep=5pt,
    showspaces=false,
    showstringspaces=false,
    showtabs=false,
    tabsize=2
}

\providecommand{\U}[1]{\protect\rule{.1in}{.1in}}

\usepackage{listings}
\usepackage{titlesec}
\titleformat{\section}
{\normalfont\Large\bfseries}{\thesection.}{1em}{}
\usepackage{fancyhdr}
\begin{document}
\date{}
\title{\Large \textbf{A Copula Based Supervised Filter for Feature Selection in Machine Learning Driven Diabetes Risk Prediction}}
\vspace{1ex}
\author{Agnideep Aich${ }^{1}$\thanks{Corresponding author: Agnideep Aich, \texttt{agnideep.aich1@louisiana.edu}, ORCID: \href{https://orcid.org/0000-0003-4432-1140}{0000-0003-4432-1140}}
 \hspace{0pt}, Md Monzur Murshed${ }^{2}$ \hspace{0pt}, Sameera Hewage${ }^{3}$\hspace{0pt} and  Amanda Mayeaux${ }^{4}$
\\ [2ex]
${ }^{1}$ Department of Mathematics, University of Louisiana at Lafayette, \\ Lafayette, LA, USA. \\  ${ }^{2}$ Department of Mathematics and Statistics, Minnesota State University, \\ Mankato, MN, USA \\ ${ }^{3}$ Department of Physical Sciences \& Mathematics, West Liberty University, \\ West Liberty, WV, USA\\ ${ }^{4}$ Department of Kinesiology, University of Louisiana at Lafayette, \\ Lafayette, LA, USA
\\ }
\date{}
\maketitle

\vspace{-20pt}
\begin{abstract}
Effective feature selection is critical for building robust and interpretable predictive models, particularly in medical applications where identifying risk factors in the most extreme patient strata is essential. Traditional methods often focus on average associations, potentially overlooking predictors whose importance is concentrated in the tails of the data distribution. In this study, we introduce a novel, computationally efficient supervised filter that leverages a Gumbel copula implied upper-tail concordance score ($\lambda_U$, a monotone transformation of Kendall’s $\tau$) to rank features by their tendency to be simultaneously extreme with the positive class. We evaluated this method against four standard baselines (Mutual Information, mRMR, ReliefF, and L1/Elastic-Net) across four classifiers on two diabetes datasets: a large-scale public health survey (CDC, $N{=}253{,}680$) and a classic clinical benchmark (PIMA, $N{=}768$). Our analysis included comprehensive statistical tests, permutation importance, and robustness checks. On the CDC dataset, our method was the fastest selector and reduced the feature space by $\approx$52\%. While this resulted in a minimal but statistically significant performance trade-off compared to using all 21 features, our filter significantly outperformed standard filters (Mutual Information, mRMR) and was statistically indistinguishable from the strong ReliefF baseline. On the PIMA dataset (8 predictors), our method's ranking produced the numerically highest ROC-AUC, despite paired DeLong tests showing no statistically significant differences versus strong baselines. PIMA thus serves as a ranking-only sanity check that our upper-tail criterion behaves sensibly in a low-dimensional clinical setting. Across both datasets, the Gumbel-$\lambda_U$ selector consistently identified clinically coherent and impactful predictors. We conclude that feature selection via upper-tail dependence is an efficient and interpretable screening approach that can complement standard feature-selection baselines in public health and clinical risk prediction.
\end{abstract}

\section*{Introduction}
\label{sec:introduction}

Diabetes mellitus is one of the biggest public health challenges of our time. The International Diabetes Federation estimates that 1 in 9 adults worldwide have diabetes, and this could rise to nearly 1 in 8 by 2050, which is about a 46\% increase \cite{IDF2025}. As more people are affected, there is a growing need for reliable and easy-to-use risk prediction tools. These tools can help clinicians and public health systems find people at high risk and offer targeted prevention.  

Machine learning has emerged as a powerful paradigm for developing such tools, leveraging large-scale health datasets like the CDC Diabetes Health Indicators \cite{CDC} survey to uncover complex patterns associated with disease risk. A critical step in building effective and trustworthy predictive models is feature selection, a process fundamental to both performance and scientific utility \cite{Guyon2003}. For a non-specialist audience, feature selection (FS) is the principled process of identifying the most informative subset of predictors from a larger pool. The goals are threefold: (i) improve predictive performance by eliminating irrelevant or noisy variables that promote overfitting; (ii) reduce cost and latency by limiting variables to measure, store, and process; and (iii) enhance interpretability to yield actionable domain insight.

Traditional filter-style selectors measure how features and labels are related overall. Pearson correlation looks at average linear relationships, while information theoretic measures like mutual information (MI)  \cite{Shannon1948, Battiti1994MI, Cover2005} capture more general, non-linear connections. However, these methods do not focus on the extremes, which are often most important for identifying high-risk cases. For example, a strong correlation between BMI and diabetes does not show what happens when BMI is at its highest levels. Our approach instead ranks features using the Gumbel copula’s upper-tail dependence coefficient ($\lambda_U$) \cite{Gumbel1960}, which highlights when extreme values of a feature and the outcome occur together.

In risk modeling, the most important information often comes from the extremes rather than the average cases. To focus on these situations, we use ideas from extreme-value theory and tail dependence \cite{Resnick1987EVT} within the copula framework. Sklar's theorem allows copulas to separate how variables are related from their individual distributions, so we can directly measure how they move together, especially in the extremes (tails) \cite{Sklar1959, Nelsen2006, Joe1997}.

We introduce and validate a supervised filter for feature selection based on the Gumbel copula’s \cite{Gumbel1960} upper-tail dependence coefficient $\lambda_U$, used here as a tail-sensitive concordance score derived from rank-based dependence. Each predictor is ranked by its $\lambda_U(X,Y)$ with the diabetes label, prioritizing features whose highest values co-occur with positive outcomes. To our knowledge, this is the first study to operationalize a copula tail-dependence coefficient as a direct, standalone criterion for supervised feature selection in clinical risk prediction.

We validate on two public datasets. On the large CDC survey, our method demonstrates a favorable trade-off between parsimony and performance. By reducing the feature set by 52\% (from 21 to 10), our Gumbel-selected model achieves a competitive ROC-AUC of 0.823 and is the fastest selector. Most importantly, it demonstrates its utility as a competitive filter by achieving statistically higher AUC scores than standard filter methods like Mutual Information and mRMR, while performing on par with the strong ReliefF baseline. 
To demonstrate external validity beyond a large, survey-style cohort (CDC), we also evaluate on the PIMA Indians Diabetes dataset, a small clinical benchmark with \textit{eight} predictors. Because PIMA contains all and only eight covariates, this experiment isolates \emph{feature ranking} rather than dimensionality reduction: all methods use the same variables. The Gumbel ranking combined with a Random Forest attains the highest ROC-AUC of 0.867, with paired DeLong tests showing no significant differences across selectors (DeLong $p>0.05$).
As such, PIMA functions as a ranking-only sanity check that our upper-tail criterion yields clinically coherent orderings without harming discrimination. Overall, our results show that focusing on the extremes can create models that are simple, easy to understand, and effective for clinical and public health use.

One might wonder why we chose the Gumbel copula and did not use a copula goodness-of-fit test. Our goal is to focus on the property of interest, not just on how well a model fits. We need a dependence measure that captures when both a predictor and the label are high at the same time. The Gumbel copula provides this, since it has nonzero upper-tail dependence and zero lower-tail dependence, giving us the coefficient we need. By contrast, the Clayton copula only shows lower-tail dependence, and the Gaussian and Frank copulas are tail-independent. Even if a goodness-of-fit test favored these other models, they would not help us understand synchrony in the upper tail. We are not fitting a full generative copula here. Instead, we use the Gumbel mapping as a scoring function that is monotonic in Kendall’s tau, so we can rank features by their upper-tail dependence. In the future, other upper-tail families, such as Joe, Student’s t, or BB1, could be compared using the same property-focused approach.

The next section reviews related work. We then present the preliminaries, introducing the methods used in this study. After that, we describe the datasets. The methodology section outlines the experimental pipeline. The results section reports our findings and includes an in-depth discussion of the medical implications. Finally, the conclusion summarizes the study and highlights directions for future research.


\section*{Related Work}
\phantomsection\label{sec:related}

Feature selection methods are usually grouped as filters, wrappers, or embedded approaches \cite{Guyon2003}. Wrappers\cite{Kohavi1997}  use a learning algorithm to evaluate feature subsets, such as genetic search or RFE, and often achieve strong results but require significant computation \cite{Holland1975, Goldberg1989}. Embedded methods combine selection with model training. Well-known examples include LASSO\cite{Tibshirani1996} and elastic-net, which encourage sparsity through regularization \cite{ ZouHastie2005, HoerlKennard1970}.

Our work introduces a filter method where features are scored directly from the data, without relying on any downstream model \cite{Blum1997}. Classic filters include mutual information (MI), which measures non-linear dependence but evaluates features one at a time \cite{Battiti1994MI}. The mRMR method addresses this by selecting features that are both highly relevant and minimally redundant \cite{Peng2005mRMR, Yu2004}. ReliefF\cite{Kononenko1994ReliefF, Kononenko1997ReliefF} focuses on local class separation by using nearest neighbors . However, these methods still mainly assess overall association. Unlike correlation-based filters (which are tail-independent), our score targets joint extremes via the Gumbel upper-tail coefficient $\lambda_U$ rather than average association.

A recent area of research uses copulas to create more fair and robust ways to measure dependence. Copula-based filters like Copula Correlation (Ccor) and Robust Copula Dependence (RCD) help reduce functional bias compared to traditional metrics \cite{Ding2015Ccor, Chang2016RCD}, which aligns with the goals of the MIC literature \cite{Reshef2011, Kinney2014}. Other advances include using copulas for dimensionality reduction \cite{Femmam2022}, entropy-based selection \cite{Ma2019, Yan2025CEFS}, hybrid filters that combine copula and information-theoretic scores to balance relevance and redundancy\cite{Lall2021}. However, most copula-based studies still focus on measuring overall dependence or redundancy between features.

Importantly, tail-dependence coefficients, which directly measure the probability of joint extremes, have not yet been used as the main supervised ranking criterion in feature selection. As seen from the literature, a Gaussian copula cannot capture tail dependence and have called for methods that use tail-dependent families. Our work addresses this need. By using the Gumbel copula's upper-tail coefficient $\lambda_U$ \cite{Joe1997, Gumbel1960}, we shift the selection focus from overall association to joint extremal behavior. This approach provides a simple, interpretable, and computationally efficient supervised filter that (i) identifies clinically meaningful predictors linked to high-risk cases, (ii) achieves performance comparable to strong baselines on both the large survey and clinical benchmark datasets, and (iii) remains fast compared to standard baselines like MI, mRMR, and ReliefF, making it practical for real-world use. 

Finally, to specifically assess the effect of tail-aware ranking when dimensionality reduction is moot, we include the 8-variable PIMA benchmark where all methods use the same variables.

Next, we present the preliminaries section.
\section*{Preliminaries}
\label{sec:preliminaries}

In this section, we introduce the main concepts and notation used in the paper, including copulas, tail-dependence coefficients, types of feature-selection methods, baseline selectors, classification models and metrics, permutation importance, and paired statistical tests.

\subsection*{Copulas}
A copula $C:[0,1]^2 \to [0,1]$ is a joint distribution with uniform marginals that captures dependence independently of the marginal distributions\cite{Nelsen2006, Joe1997}. By Sklar's theorem \cite{Sklar1959}, for any bivariate cdf $H$ with continuous marginals $F_X, F_Y$ there exists a unique copula $C$ such that

\begin{align}
H(x,y) = C(F_X(x), F_Y(y)).
\end{align}

Consequently, if we convert data to pseudo-observations

\begin{align}
U = \frac{\text{rank}(X)}{n+1}, \quad V = \frac{\text{rank}(Y)}{n+1},
\end{align}

any measure computed from $C$ depends only on the dependence structure, not on the marginal scales. Copulas are organized in families (e.g., elliptical, Archimedean); we will later specialize to the Gumbel \cite{Gumbel1960} copula, which is part of the Archimedean family, because it exhibits positive upper-tail clustering, which is the signal we target for feature screening.

\subsection*{Tail Dependence in Copulas}
``Tail dependence'' expresses the probability that one variable is extreme given the other is extreme. Tail dependence is symmetric in the variables (i.e., invariant to swapping $U$ and $V$ in the bivariate case). For the upper tail, the coefficient $\lambda_U \in [0,1]$ is defined as
\begin{align}
\lambda_U = \lim_{q \to 1^-} \Pr(V > q \mid U > q).
\end{align}
Intuitively, $\lambda_U$ is the asymptotic co-occurrence probability of simultaneous highs. Values near 1 indicate strong synchronization at extreme levels; values near 0 indicate little to no joint extremal behavior. The lower-tail analogue $\lambda_L = \lim_{q \to 0^+} \Pr(V \leq q \mid U \leq q)$ exists but is not used in this work.

\noindent\textbf{Scope.} Tail-dependence concepts are most naturally defined for continuous margins; in this paper, we use the Gumbel mapping as a rank-based, tail-sensitive concordance score (a monotone transformation of Kendall’s $\tau$) for feature ranking, rather than as a generative copula tail parameter. Throughout the paper, references to $\lambda_U$ (upper-tail dependence) are to this Gumbel-implied, rank-based concordance score derived from Kendall’s $\tau$, used for ranking rather than for generative copula modeling.

\paragraph{Feature Ranking via $\lambda_U$.}
For each feature $X_j$ and the binary label $Y$, we form pseudo-observations $U_j, V$ from ranks, estimate Kendall's $\tau(U_j, V)$, and (under the Gumbel family) map $\tau$ to the copula parameter and then to $\lambda_U$. The resulting $\lambda_U^{(j)}$ serves as a feature score: features with larger $\lambda_U^{(j)}$ exhibit stronger joint extremes with the outcome and are ranked higher.

Throughout, ``extremes” and “movement in the tails” refer to the same concept: the synchronized co-occurrence of upper-tail values of a feature and the label, quantified by the Gumbel upper-tail dependence coefficient $\lambda_U$.

\subsection*{Gumbel copula}
The Gumbel (Gumbel–Hougaard) \cite{Gumbel1960, Hougaard1986} copula is an Archimedean copula with generator
\begin{align}
C_{\theta}(u,v)
= \exp\!\left\{-\Big[(-\log u)^{\theta}+(-\log v)^{\theta}\Big]^{1/\theta}\right\},\qquad (u,v)\in(0,1)^2.
\end{align}
Gumbel exhibits upper–tail but not lower–tail dependence:
\begin{align}
\lambda_U(\theta)=2-2^{\,1/\theta}, \qquad \lambda_L(\theta)=0.
\end{align}
Kendall’s tau for Gumbel is
\begin{align}
\tau(\theta)=1-\frac{1}{\theta}\quad\Longleftrightarrow\quad
\theta=\frac{1}{1-\tau}\;\;(\tau>0).
\end{align}
Thus, \(\tau\in[0,1)\); \(\tau=0\) corresponds to independence (\(\theta=1\)). In our scoring, for each feature $X_j$ we estimate Kendall’s \(\hat\tau(X_j,Y)\) on pseudo-observations; if \(\hat\tau>0\) we map
\[
\hat\theta_j=\frac{1}{1-\hat\tau},\qquad
\hat\lambda_U^{(j)}=2-2^{\,1/\hat\theta_j},
\]
and if \(\hat\tau\le 0\) we set \(\hat\lambda_U^{(j)}:=0\) (no positive upper-tail co-occurrence under Gumbel).
\smallskip

We use a property-first approach and require a dependence measure with nonzero \(\lambda_U\). The Gumbel copula meets this requirement, since \(\lambda_U>0\) and \(\lambda_L=0\). In contrast, the Clayton copula only has lower-tail dependence (\(\lambda_U=0\)), and the Gaussian and Frank copulas are tail-independent (\(\lambda_U=\lambda_L=0\)). Even if a goodness-of-fit test favored those other families, they would not address our question about upper-tail synchrony. For this reason, we use the Gumbel mapping as a monotone transform of \(\tau\) to rank features by upper-tail dependence. We are not claiming a full generative copula fit. While the classical upper-tail dependence coefficient is defined asymptotically for continuous margins, here it is used as a monotone transformation of Kendall’s \(\tau\) to construct a tail-sensitive concordance score for ranking; we do not claim estimation of asymptotic tail probabilities or a fully specified generative copula model.

\paragraph{Computational cost.}
Let $n$ be the number of samples and $d$ the number of features. Our selector does \emph{not} fit a model per feature. For each $X_j$ we compute pseudo-observation ranks ($O(n\log n)$; $V$ is ranked once), compute Kendall’s $\tau(U_j,V)$ via an $O(n\log n)$ exact algorithm~\cite{Knight1966}, and apply the closed-form maps $\theta=1/(1-\tau)$ and $\lambda_U=2-2^{1/\theta}$ for the Gumbel copula~\cite{Joe1997,Nelsen2006}. Thus the total selection cost is $O(d\,n\log n)$ (and $O(d\,n)$ if an approximate/linear-time $\tau$ is used), with much smaller constants than wrappers/embedded methods that fit models. Empirically (Results, Tables~\ref{tab:cdc_fs_time}, \ref{tab:pima_fs_time}), Gumbel--$\lambda_U$ is the fastest selector on CDC and comparable to MI/mRMR on PIMA. Pearson correlation is cheaper but tail-agnostic and cannot target joint extremes (e.g., Gaussian/Frank are tail independent).

\subsection*{Families of Feature Selection Methods}
Feature selection (FS) methods are commonly grouped as:
\begin{itemize}
\item \textbf{Filter} methods \cite{Guyon2003}: model-agnostic scoring of features using data statistics (e.g., MI, correlation, ReliefF, our $\lambda_U$ score). They do not fit the predictor during selection and are computationally efficient.
\item \textbf{Wrapper} methods \cite{Blum1997}: search over subsets guided by a predictive model (e.g., forward selection, RFE); usually costly.
\item \textbf{Embedded} methods \cite{Guyon2003}: selection occurs within model training (e.g., L1/elastic-net logistic regression, tree-based splits).
\item \textbf{Hybrid} methods \cite{Hsu2011}: combination of filter and wrapper methods. The filter approach selects the candidate feature set from the original feature set and refines the candidate feature set by the wrapper methods. It brings together the strengths of both methods.
\end{itemize}
Feature selection approaches can also be categorized according to selection criteria into four groups \cite{Dhal2022}: statistical measure-based, information theory-based, similarity measure-based, and sparse learning-based.

Our proposed Gumbel-$\lambda_U$ procedure is a supervised filter method. In comparisons, MI, mRMR, and ReliefF are also filters; L1/Elastic-Net is an embedded selector. Benchmarking the proposed method against wrapper and hybrid methods will be left for future work.

\paragraph{Mutual Information (MI):} 
MI \cite{Shannon1948, Battiti1994MI} is a filter method that measures how far the joint behavior of two variables deviates from what would be expected if they were independent. If $X$ and $Y$ were independent, the joint distribution would factor as $p(x,y) = p(x)p(y)$. MI quantifies how the actual joint relationship between two variables differs from this independence baseline; a greater deviation indicates a stronger dependence, whether linear or non-linear. 

\begin{align}
I(X;Y) = E\!\left[\log \frac{p(X,Y)}{p(X)p(Y)}\right] 
= \sum_x \sum_y p(x,y) \log \frac{p(x,y)}{p(x)p(y)}.
\end{align}

$I(X;Y) \geq 0$, and $I(X;Y) = 0$ if and only if $X$ and $Y$ are independent.
In feature selection, MI is used to rank features by their dependence with the target variable, typically by computing $I(X_j;Y)$ for each feature $X_j$.

\paragraph{mRMR (Minimum-Redundancy, Maximum-Relevance):} 
mRMR \cite{Peng2005mRMR, Yu2004} is a filter method that builds a feature set step-by-step. At each step, it prefers a feature that is (i) relevant to the label (informative on its own) and (ii) not redundant with features already chosen (brings new information). A common score is the difference between ``MI with $Y$'' and the ``average similarity to selected features''. In practice, similarity is often measured by absolute correlation for computational efficiency. In feature selection, mRMR builds the feature set greedily by adding the feature with the highest ``relevance -- redundancy'' score, i.e., maximizing $I(X_j;Y)$ minus its average correlation with already-selected features, and stops after selecting $k$ features.

\paragraph{ReliefF:} 
ReliefF \cite{Kononenko1994ReliefF, Kononenko1997ReliefF} is a filter method that takes a nearest-neighbor view of feature usefulness. For each data point, it finds the closest point in the same class (a ``hit'') and the closest point in the other class (a ``miss''). A feature receives a higher score if, along that feature, the point is close to its hit but far from its miss, meaning the feature helps separate classes in local neighborhoods. This approach captures non-linear structure and some feature interactions. In feature selection, ReliefF evaluates many sampled points by comparing each feature’s average difference to nearest neighbors of the other class with that of the same class. Features are then ranked or selected based on the resulting average score. Larger scores indicate more discriminative features.

\paragraph{L1 / Elastic-Net Logistic Regression (L1EN):} 




L1EN \cite{ZouHastie2005, Tibshirani1996, HoerlKennard1970} is an embedded method that uses both $L1$ and $L2$ regularization penalties, with the emphasis on $L1$ for feature selection. The $L1$ (lasso) penalty, $\|\beta\|_1 = \sum_j |\beta_j|$, tends to set many coefficients exactly to zero, producing a sparse model where selected features correspond to the non-zero coefficients. The $L2$ (ridge) penalty, $\|\beta\|_2^2 = \sum_j \beta_j^2$, shrinks coefficients toward zero but rarely makes them exactly zero, stabilizing estimates when features are correlated. Elastic Net combines both penalties, $\alpha \|\beta\|_1 + \frac{1-\alpha}{2} \|\beta\|_2^2$, providing sparsity from $L1$ and stability from $L2$. When a penalized logistic regression is fitted with these penalties, the $L1$ component determines which coefficients are exactly zero, with a non-zero coefficient $\beta_j$ indicating that feature $j$ is retained and a zero coefficient effectively removing it. Feature selection can then be performed by ranking features according to the absolute values of their coefficients, $|\beta_j|$, with larger values indicating greater importance, and selecting the top $k$ features based on this ranking.

\subsection*{Classifiers Used for Evaluation }

\paragraph{Random forest (RF):} RF \cite{Breiman2001} creates many decision trees using different random samples of the data and by choosing random subsets of features for each split. The final prediction is made by combining the results from all the trees, either by voting or averaging. RF is effective for both classification and regression tasks. It reduces variance through averaging, handles non-linearities and feature interactions, and performs well on tabular data with mixed signals.



\paragraph{Gradient Boosting (GB):} GB \cite{Friedman2001} constructs a sequence of trees, with each new tree trained to correct the errors of the preceding ones. The final model aggregates the predictions of all trees. GB is applicable to both classification and regression tasks. It offers strong predictive performance on tabular data, with fine control over overfitting through the learning rate and tree depth.


\paragraph{Extreme Gradient Boosting (XGB):} XGB \cite{Chen2016} is a high-performance, regularized implementation of gradient boosting that incorporates system-level optimizations such as parallelism and sparsity-aware learning. It is typically fast and accurate on medium to large tabular datasets, and its regularization mechanisms help mitigate issues when features are correlated.


\paragraph{Logistic Regression (LR):} LR \cite{Cox1958}, despite its name, is primarily used for classification tasks with two or more classes. It computes a linear combination of the features and applies a logistic or softmax function to convert this score into class probabilities. LR is simple and interpretable, often provides well-calibrated probabilities, and with regularization, it resists overfitting and can serve as an embedded feature selector.

\subsection*{Performance Metrics}

We use quantitative performance metrics to evaluate how well machine learning models work. These metrics help us compare different models and choose the best one for a specific task. In classification problems, the most common metrics are accuracy, precision, recall, F1 score, and area under the ROC curve (AUC). 

\paragraph{Accuracy:} Accuracy measures the proportion of correctly classified observations among all observations:
\begin{align}
\text{Accuracy} = \frac{TP + TN}{TP + TN + FP + FN},
\end{align}
where $TP$ = true positives, $TN$ = true negatives, $FP$ = false positives, and $FN$ = false negatives.

Accuracy is intuitive and easy to compute. In imbalanced datasets, high accuracy may be misleading because the model can achieve high scores simply by predicting the majority class.

\paragraph{Precision:}

Precision measures the proportion of predicted positives that are truly positive:
\begin{align}
\text{Precision} = \frac{TP}{TP + FP}.
\end{align}
Precision is useful when false positives are costly (e.g., diagnosing a disease when the patient is healthy).

\paragraph{Recall:}

Recall (also called sensitivity or true positive rate) measures the proportion of actual positives that were correctly identified:
\begin{align}
\text{Recall} = \frac{TP}{TP + FN}.
\end{align}
Recall is useful when false negatives are costly (e.g., failing to detect a disease in a patient who has it).

\paragraph{F1 Score:}

The F1 score combines precision and recall into a single metric, defined as the harmonic mean:
\begin{align}
F1 = 2 \cdot \frac{\text{Precision} \cdot \text{Recall}}{\text{Precision} + \text{Recall}}.
\end{align}

$F1$ balances precision and recall. It is especially valuable when dealing with imbalanced data, where accuracy alone can be misleading.

\paragraph{ROC and AUC (ROC-AUC):} The ROC (Receiver Operating Characteristic) is a curve: it plots the true positive rate (TPR, i.e., recall) against the false positive rate (FPR) as the decision threshold sweeps from 1 to 0. Here
\begin{align}
\text{TPR}=\frac{\mathrm{TP}}{\mathrm{TP}+\mathrm{FN}},\qquad \text{FPR}=\frac{\mathrm{FP}}{\mathrm{FP}+\mathrm{TN}}.
\end{align}
The AUC is a number: the Area Under that ROC curve, with $0\le \mathrm{AUC}\le 1$. AUC summarizes threshold-free discrimination: it equals the probability that a randomly chosen positive instance receives a higher model score than a randomly chosen negative instance,
\begin{align}
\mathrm{AUC} = \Pr(s^+ > s^-) + \frac{1}{2}\Pr(s^+ = s^-).
\end{align}

Thus, ROC-AUC jointly refers to the ROC curve and its area. Values near 1 indicate strong ranking ability; 0.5 corresponds to random guessing. Because it does not depend on a single threshold (and is less sensitive to class imbalance than accuracy), ROC-AUC is our primary metric.

\subsection*{Permutation Importance (Model-Agnostic Interpretation)}
For a fitted model $f$ and test data $(X_{\text{test}}, y_{\text{test}})$, the permutation importance of feature $j$ measures the drop in performance when the $j$-th column is randomly permuted:
\begin{align}
I_j = E[\text{AUC}(f; X_{\text{test}}, y_{\text{test}})] - E[\text{AUC}(f; X_{\text{test}}^{(j)}, y_{\text{test}})].
\end{align}

We report the mean and standard deviation over $n_{\text{repeats}}$ permutations (default $500$), using ROC-AUC as the scorer. We compute permutation importance for the best Gumbel configuration (GB on CDC; RF on PIMA).

\subsection*{Paired Statistical Tests}
To compare models fairly on the same test split, we use two paired tests:

\paragraph{DeLong's test for AUC.} DeLong's test \cite{DeLong1988} is a non-parametric method for comparing the Area Under the Curve (AUC) of two correlated ROC curves, which arise when two different models are evaluated on the same test set. The test accounts for the covariance between the models' predictions to provide a more accurate variance estimate. The null and alternative hypotheses are:
\begin{itemize}
    \item \textbf{H$_0$ (Null Hypothesis):} The AUC of Model A is equal to the AUC of Model B.
    \item \textbf{H$_1$ (Alternative Hypothesis):} The AUC of Model A is not equal to the AUC of Model B.
\end{itemize}
A high p-value (e.g., $p > 0.05$) suggests that any observed difference in AUC is not statistically significant. It does not give the direction. Direction comes from the numerical AUCs themselves.

\paragraph{McNemar's test for error discordance.} McNemar's test \cite{McNemar1947} is a non-parametric test used to compare the classification errors of two models. It operates on a 2x2 contingency table that cross-classifies the instances based on whether each model's prediction was correct or incorrect. The test determines if there is a significant difference in the models' disagreement rates. We use the continuity-corrected version. The null and alternative hypotheses are:
\begin{itemize}
    \item \textbf{H$_0$ (Null Hypothesis):} The two models have the same error rate (i.e., their disagreements are symmetric).
    \item \textbf{H$_1$ (Alternative Hypothesis):} The two models have different error rates.
\end{itemize}
A low p-value (e.g., p < 0.05) indicates that one model is making significantly more errors on a different set of instances than the other, suggesting a meaningful difference in their predictive behavior.

In the next section, we describe the two datasets used in our study.


\section*{Dataset Description}
\label{sec:dataset}

In this section, we describe the two datasets we have used in this study.

\subsection*{Dataset Description (CDC Diabetes Health Indicators)}
The first dataset used in this study is the CDC Diabetes Health Indicators dataset, comprising \(253,680\) individuals and all \(21\) original features. These span demographic, lifestyle, and health-related characteristics (Table~\ref{tab:summary_features}), providing a comprehensive basis for diabetes risk analysis.

The dataset was accessed via the UCI Machine Learning Repository \verb|CDC Diabetes Health Indicators| \cite{CDC} using the \texttt{ucimlrepo} Python library. All features were validated to contain no missing or constant values.

\textbf{Target Variable:} The dataset includes a binary label \texttt{Diabetes\_binary}, where 0 means no diabetes and 1 means prediabetes or diabetes. Our study treats diabetes risk prediction as a binary classification problem, in line with the repository label.
This definition enables focused evaluation of risk factors and classification performance.

\textbf{Features:} The dataset includes \(21\) features spanning five categories: demographic variables, lifestyle factors, health indicators, general and mental health indicators, and healthcare access. These are summarized in Table~\ref{tab:summary_features}, highlighting the comprehensive scope of the dataset.

\begin{table}[ht]
\centering
\caption{Summary of features in the CDC Diabetes Health Indicators dataset. Features are grouped into thematic categories relevant to risk prediction and healthcare modeling.}
\label{tab:summary_features}
\begin{tabularx}{\textwidth}{@{} l X @{}}
\toprule
\textbf{Category} & \textbf{Features} \\
\midrule
Demographic Variables & Age, Sex, Income, Education Level \\
Lifestyle Factors & Smoking Status, Physical Activity, Fruit and Vegetable Consumption, Alcohol Consumption \\
Health Indicators & High Blood Pressure, High Cholesterol, Coronary Heart Disease or Heart Attack, Stroke History \\
General and Mental Health Indicators & Self-Reported General Health, Days of Poor Mental Health, Days of Poor Physical Health \\
Healthcare Access & Health Insurance Coverage, Instances of Being Unable to See a Doctor Due to Cost \\
\bottomrule
\end{tabularx}
\end{table}

\textbf{Data Characteristics:} The dataset is clean, with no missing or constant values. The features are well-defined, encompassing diverse factors relevant to the study's objective.

\textbf{Purpose:} Created by the Centers for Disease Control and Prevention (CDC), this dataset is specifically designed to examine the interplay between lifestyle, healthcare accessibility, and diabetes prevalence in the US population. The insights derived from this dataset have broad implications for public health strategies and intervention design.

\subsection*{Dataset Description (PIMA)} \label{sec:dataset-pima}
We additionally evaluate our method on a second dataset namely the \textit{Pima Indians Diabetes Database}, a widely used clinical benchmark originating from the National Institute of Diabetes and Digestive and Kidney Diseases (NIDDK) and described by Smith \textit{et al.}~\cite{Smith1988}. The dataset comprises 768 adult female patients (age $\geq$ 21) of Pima Indian heritage, with \textbf{eight} diagnostic predictors and a binary diabetes label \texttt{Outcome}. A curated copy of this dataset was obtained from Kaggle.

\textbf{Target Variable:} The dataset includes a binary label \texttt{Outcome} (0 = no diabetes, 1 = diabetes). We use this label as provided by the repository—no relabeling or class collapsing was performed. Our study treats diabetes risk prediction as a binary classification problem, in line with the repository label. This definition enables focused evaluation of risk factors and classification performance.

\textbf{Features:} The eight predictors are: \emph{Pregnancies}, \emph{Glucose}, \emph{BloodPressure}, \emph{SkinThickness}, \emph{Insulin}, \emph{BMI}, \emph{DiabetesPedigreeFunction}, and \emph{Age}. For completeness, Table~\ref{tab:pima_features} lists the variables and brief descriptions.

\begin{table}[ht]
\centering
\caption{\textbf{Summary of features in the PIMA dataset.} Brief one-line descriptions of each diagnostic predictor.}
\label{tab:pima_features}
\begin{tabularx}{\textwidth}{@{} l X @{}}
\toprule
\textbf{Feature} & \textbf{Description} \\
\midrule
Pregnancies & Number of pregnancies. \\
Glucose & 2-hour plasma glucose concentration (oral glucose tolerance test). \\
BloodPressure & Diastolic blood pressure (mm Hg). \\
SkinThickness & Triceps skinfold thickness (mm). \\
Insulin & 2-hour serum insulin (mu U/ml). \\
BMI & Body mass index $\left(\frac{\text{kg}}{\text{m}^2}\right)$. \\
DiabetesPedigreeFunction & Proxy for hereditary risk (family history signal). \\
Age & Age in years. \\
\bottomrule
\end{tabularx}
\end{table}

\textbf{Data Characteristics:} The working CSV (768 $\times$ 9) contains no NA/missing values. As is standard for this dataset, several physiological fields may contain literal zeros (e.g., \emph{Glucose}, \emph{BloodPressure}, \emph{SkinThickness}, \emph{Insulin}, \emph{BMI}); these zeros are present in the released files and are not NA placeholders. (Handling of such values, if any, is documented in the Methodology.)

\textbf{Provenance and reference:} The dataset traces to NIDDK and the study of Smith et al.~\cite{Smith1988}; we cite the original source and also provide the Kaggle page link (in the additional information section below) that was used to obtain the CSV.

\noindent\textbf{Note on scope.}
Since PIMA has exactly eight predictors, all methods ultimately use the same variables. We report rankings only for interpretability; statistical tests (paired DeLong and McNemar) verify that any observed performance differences are not significant.

In the next section, we describe the methodology. This includes Gumbel-$\lambda_U$-based feature selection (Alg.~\ref{alg:gumbel_feature_selection}), benchmark comparators, model fitting, threshold selection, and evaluation.

\section*{Methodology}
\label{sec:methodology}

For tree ensembles (RF/GB/XGB), predicted probabilities were calibrated via 5-fold isotonic regression  \cite{Zadrozny2002, NiculescuMizil2005, Guo2017}. Logistic regression was left uncalibrated. Calibration was fit on the training folds only and applied to the validation/test sets. We calibrated the tree-ensemble probabilities because uncalibrated trees are typically miscalibrated (often overconfident), whereas regularized logistic regression is generally close to calibrated; therefore, we left LR uncalibrated. Logistic regression used standardized features, while tree-based models used raw inputs. Decision thresholds were selected on the validation set to maximize F1 and then held fixed for test evaluation. All experiments used a fixed random seed and identical train/validation/test splits.

We standardize inputs only where the downstream method requires distance or coefficient comparability:(i) LR is trained on standardized predictors (scaler fit on train only and applied to val/test) to stabilize the optimizer and penalty. (ii) ReliefF relies on Euclidean nearest neighbors, so we z-score each feature on the training split only before its hit/miss computations and apply the learned mean/SD to val/test. (iii) L1EN also operates on features that have been scaled to have a zero mean and unit variance. By contrast, (iv) tree ensembles (RF/GB/XGB) are scale-invariant and use raw inputs; and (v) the MI and mRMR filters are used without scaling: MI here is univariate $I(X_j;Y)$, so linear rescaling of $X_j$ does not change the information content, and the mRMR redundancy term uses Pearson correlation, which is itself scale-invariant. All scalers are train-only to avoid leakage.

\paragraph{Research question:} Which predictors exhibit upper-tail co-occurrence with diabetes status, i.e., when a feature is high, how likely is the outcome to be positive and does selecting features by this property improve predictive performance over standard baselines?

For each feature $X_j$, we compute the Gumbel upper–tail dependence score $\lambda_U(X_j,Y)$ from rank-based pseudo-observations and \textit{rank features by} $\lambda_U$. We say a feature shows a positive upper-tail association when the 95\% bootstrap percentile confidence interval for $\lambda_U(X_j,Y)$, computed from $B=1000$ bootstrap resamples of the \emph{training split} (with replacement), lies strictly above $0$. We keep the top-$k$ features and train models on that set. To judge usefulness, we compare these models with baselines (MI, mRMR, ReliefF, and L1/Elastic-Net) on the same test split, using ROC-AUC as the primary metric (with PR–AUC, accuracy, recall, and F1 as secondary metrics). For ROC-AUC, we also report 95\% bootstrap percentile confidence intervals on the test set using $B=1000$ resamples of test indices. For formal paired comparisons we use DeLong’s test for ROC-AUC and McNemar’s test for error profiles.

Guided by the research question above, we rank features by $\lambda_U$, fix $k$ per dataset (CDC: $k{=}10$; PIMA: $k{=}8$), and evaluate the resulting sets alongside baselines (MI, mRMR, ReliefF, L1EN) using four classifiers (RF, XGB, LR, GB) with class-balanced training and probability calibration. We report Accuracy, Precision, Recall, F1, and ROC-AUC; we compute permutation importance on the held-out test set; and we use DeLong and McNemar tests for paired comparison. Computationally, computing Kendall’s $\tau$ per feature with a closed-form transform yields an $O(d\,n\log n)$ selector with no model fitting in the selection loop.

We study two public datasets: CDC Diabetes Health Indicators dataset and the PIMA Indians Diabetes dataset. For CDC, we use all 21 features as provided. The dataset has no missing values. For PIMA, we use the common practice of treating zeros as missing for five physiological variables: Glucose, BloodPressure, SkinThickness, Insulin, and BMI. In the training split only, we replace zeros in these columns with NaN and fit a median imputer. We then apply the learned medians to both the training and test splits. 

We split each dataset once into training and test sets using an 80/20 stratified split. From the training portion, we hold out a validation subset, which is 20\% of the training data. This subset is used only to select decision thresholds; we never tune models on the test set. We address class imbalance in two ways. First, when training logistic regression, random forests, and gradient boosting, we set class weights to ``balanced''. For XGBoost, we calculate \texttt{scale\_pos\_weight} as the ratio of negatives to positives in the training data. Second, instead of fixing the probability cutoff at 0.5, we choose the operating threshold on the validation data by scanning a fine grid of probability quantiles and selecting the cutoff that maximizes the F1 score. This approach keeps operating points comparable across different class prevalences and reduces the risk that an imbalanced base rate affects the differences in accuracy.

Our feature selection stage compares four standard baselines with the proposed Gumbel-$\lambda_U$ criterion, all calculated on the training data. 

\begin{enumerate}

\item The mutual information (MI) baseline ranks each feature using \texttt{mutual\_info\_classif}, which employs a $k$-nearest-neighbors entropy estimator with the library's default settings (effectively $k = 3$ and continuous features) under our fixed random seed. Features are sorted by the resulting $\hat{I}(X_j; Y)$. 

\item The elastic-net baseline fits a logistic regression using the SAGA solver on standardized predictors. It runs for up to 4,000 iterations with \texttt{class\_weight=``balanced''} and ranks features by the absolute coefficient magnitudes at the end.  

\item The mRMR implementation is a simple greedy method that selects the first feature with the highest mutual information to the label (using the same estimator as before). It then adds the feature that maximizes ``relevance minus redundancy'', where redundancy is the mean absolute Pearson correlation with the already chosen features. 

\item ReliefF uses the standard binary method: we z-score each feature and then find ten nearest ``hits'' (same class) and ten nearest ``misses'' (opposite class) for every training instance using Euclidean distance. We accumulate a score equal to the average miss-distance minus the average hit-distance across all rows. Larger values indicate stronger class separation.
\end{enumerate}

Our method is a supervised filter feature selector. It is supervised because the score for each feature $X_j$ uses the label $Y$ (via $\lambda_U(X_j, Y)$). It is a filter because features are scored and ranked without fitting any predictive model during selection; the score depends only on the data's dependence structure (via pseudo-observations and the Gumbel mapping), not on a particular classifier. Because the score is computed from pseudo-observations (ranks), it is invariant to monotone rescalings of $X_j$ and does not assume common measurement units. By contrast, wrappers (e.g., forward selection, RFE) repeatedly train a model to evaluate subsets, and embedded methods (e.g., L1/Elastic-Net) perform selection inside model training.

Our method focuses on upper-tail dependence between each feature and the outcome (response/label) variable. We convert both variables into pseudo-observations using their ranks. We map each variable $x$ to pseudo-observations using equation~\eqref{eq:rank} and we use average ranks for ties. The same process applies to the binary label $y$.
\begin{align}
u_i = \frac{\operatorname{rank}(x_i)}{n+1}, \qquad i = 1, \ldots, n,
\label{eq:rank}
\end{align}

 We calculate Kendall's $\tau$ for each feature against the label. Then, we map $\tau$ to the Gumbel copula parameter with the formula $\hat{\theta} = 1/(1-\tau)$ whenever $\tau > 0$. After that, we convert $\hat{\theta}$ into the upper-tail dependence coefficient $\lambda_U = 2 - 2^{1/\hat{\theta}}$. We rank the features based on $\lambda_U$. To quantify uncertainty in the tail-dependence scores, we bootstrap the training split $B=1000$ times with replacement, recompute $\lambda_U$ for each resample, and report the bootstrap mean and the 2.5th–97.5th percentile interval.

Because Gumbel models positive upper-tail clustering, we set $\lambda_U^{(j)}{=}0$ when $\tau(U_j,V)\le 0$. Intuitively, a non-positive $\tau$ indicates no positive concordance (or an inverse association), which does not constitute upper-tail co-occurrence; such features are therefore de-prioritized by design. In CDC, several variables had $\tau<0$ (e.g., PhysActivity, Fruits, Veggies, HvyAlcoholConsump, Education, Income) and so received $\lambda_U{=}0$. In PIMA, all eight predictors had $\tau>0$, so all received strictly positive $\lambda_U$. If protective (lower-tail) signals are of interest, a lower-tail coefficient (e.g., $\lambda_L$ via Clayton or a symmetric family with both tails) could be used, which we leave for future work.

For the CDC dataset, we chose the number of features via a small sweep,
\(k \in \{3,5,7,10\}\), and fixed \(k=10\) for all CDC analyses. 
For PIMA, no \(k\)-sweep was performed: the dataset has exactly eight predictors, so all methods use the same eight variables and our criterion serves as a ranking-only check. Models were trained on the training split, the operating threshold was selected on a validation split using F1, and test performance was summarized by ROC-AUC (primary) and related metrics. See Aich (2026)~\cite{Aich2025} for additional CDC sweep details.

We evaluate a diverse set of classifiers on each feature set. Logistic regression (LR) is used with balanced class weights, while random forests (RF) are also applied with balanced class weights. Gradient boosting (GB) is implemented using the standard \texttt{scikit-learn} version with a fixed random seed, and XGBoost (XGB) is employed with its standard setup, handling class weights as described below.

To address class imbalance during model training, we incorporate specific weighting strategies. For LR and RF, the \texttt{class\_weight=``balanced''} parameter is used, which automatically adjusts weights to be inversely proportional to the class frequencies present in the training data. For GB, we emulate class balancing by passing per-sample weights proportional to the inverse class frequencies. For XGB, we set the \texttt{scale\_pos\_weight} parameter equal to the ratio of the number of negative samples to positive samples.

For the tree-based models (RF, GB, XGB), we use raw feature values, while for logistic regression (LR), we standardize predictors by fitting a scaler on the training data. LR is not calibrated, whereas RF, GB, and XGB models use isotonic calibration. Our evaluation pipeline then follows a careful two-stage fit to prevent data leakage:

\begin{enumerate}
    \item \textbf{Stage 1 (Threshold Selection):} We first train an initial model on a smaller training subset (80\% of the full training data), applying the class balancing strategies defined above. For RF, GB, and XGB models, we also perform five-fold isotonic calibration on this subset. This model is then used to generate probability scores on a held-out validation set (the remaining 20\% of the training data), from which we select the optimal probability threshold that maximizes the F1-score.
    \item \textbf{Stage 2 (Final Fit and Testing):} With the threshold now fixed, we train a final model on the entire training set, which combines the smaller training and validation subsets. The model architecture, balancing, and calibration strategies remain identical to Stage 1, but the feature scalers and probability calibrators are re-fit on this larger dataset. This final model is then evaluated on the held-out test set, using the pre-determined threshold to assign class labels.
\end{enumerate}

This two-stage process ensures the test set remains completely unseen and that the operating threshold is selected without using the final model's training data. We report ROC-AUC, accuracy, precision, recall, and F1-score on the test set.

To check if the observed differences are statistically significant, we run paired tests on the test set based on the Gumbel selection. For ROC-AUC, we use the DeLong test to compare the best model trained on Gumbel-selected features with the same model trained on each baseline feature set. This gives us a p-value for each paired AUC difference. At the same time, we compare binary correctness vectors between the anchored model and alternatives using McNemar's test with a continuity correction. This test determines whether the two classifiers disagree symmetrically on test instances.

We evaluate the model's behavior using permutation importance on the held-out test inputs. First, we rank all (feature set, model) combinations by ROC-AUC. We select the single best pair, fit its model to the entire training data as described earlier, and calculate permutation importance by randomly permuting each feature 500 times. We measure the average drop in ROC-AUC; we apply the same preprocessing and calibration used during training before scoring. Next, we repeat this process specifically for the Gumbel top-$k$ feature set, using the best-performing model within Gumbel, again with 500 permutations per feature. These two approaches allow us to verify that the selected variables are significant for generalization performance.

\paragraph{Ordering for ``All'' vs.\ ranked sets.}
For CDC, ``All'' uses every feature. For PIMA, all methods use the same eight predictors; column order has no effect for LR/RF/GB/XGB. We include rankings only for interpretability; as expected, AUCs are statistically indistinguishable across selectors (DeLong p > 0.05).

Finally, we examine how robust our model is under light, controlled changes to the training and test inputs. We use the gradient boosting classifier with the Gumbel top-$k$ features. Starting with a baseline run using the original data, we look at three stressors. 

\begin{enumerate}

\item The first is label noise, where we randomly flip 5\% of the training labels. 

\item The second is feature noise, where we add zero-mean Gaussian noise equal to 10\% of each feature's training-set standard deviation to both the training and test sets. 

\item The third is MCAR missingness (missing completely at random), where we randomly set 10\%, 20\%, and 30\% of the entries to missing per feature and imputed them using the training median of that feature. In this paper, we report results only for the 10\% missingness scenario, while the 20\% and 30\% results are provided in the dissertation\cite{Aich2025}.

\end{enumerate}

We evaluate each change using the same training process, and we record ROC-AUC scores on the unchanged test labels.

\subsection*{Algorithm}
\label{lab:algorithm}
In this section, we present two related algorithms. \textbf{Algorithm~\ref{alg:gumbel_feature_selection}} outlines our method for selecting features based on the Gumbel copula's upper-tail dependence. Each feature is turned into pseudo-observations and linked to the label using Kendall's $\tau$. These are then mapped to the Gumbel parameter $\theta$ and ranked by $\lambda_U$. We select the top-$k$ features (CDC: $k = 10$; PIMA: $k = 8$) for the predictive pipeline. \textbf{Algorithm~\ref{alg:full_pipeline}} summarizes the complete experimental protocol. This includes stratified splits, validation-based thresholding using the F1 score, calibrated models that handle class imbalance, test-set metrics and curves, permutation importance measured by the ROC-AUC scorer with \texttt{n\_repeats=500}, bootstrap confidence intervals for $\lambda_U$ ($B = 1000$) and robustness checks, and significance testing. Together, these algorithms match our implementation and the results presented in the paper.
\begin{algorithm}[ht]
\caption{Feature Selection via Gumbel Copula Upper–Tail Dependence}
\label{alg:gumbel_feature_selection}
\begin{algorithmic}[1]
\Require Dataset $\mathcal{D}=\{(\mathbf{x}_i,y_i)\}_{i=1}^n$ with $\mathbf{x}_i\in\mathbb{R}^d$, $y_i\in\{0,1\}$; desired $k$
\Ensure Top-$k$ features ranked by upper–tail dependence $\lambda_U$

\State \textbf{Pseudo–observations.} For each feature $X_j$ and the label $Y$, define
\[
U_j \gets \frac{\mathrm{rank}(X_j)}{n+1}, \qquad V \gets \frac{\mathrm{rank}(Y)}{n+1}.
\]
\State \textbf{Association (Kendall’s $\tau$).} For each $j=1,\dots,d$, compute Kendall’s $\tau_j=\tau(U_j,V)$.
\State \textbf{Gumbel parameter.} Map $\tau_j$ to the Gumbel copula parameter
\[
\theta_j \gets 
\begin{cases}
\frac{1}{1-\tau_j}, & \text{if }\tau_j>0,\\
\text{undefined}, & \text{if }\tau_j \le 0,
\end{cases}
\]
\State \textbf{Upper–tail dependence.} For each $j$, set
\[
\lambda_U^{(j)} \gets 
\begin{cases}
2 - 2^{\,1/\theta_j}, & \text{if }\theta_j\ge 1,\\
0, & \text{otherwise}.
\end{cases}
\]
\State \textbf{Rank \& select.} Sort features by $\lambda_U^{(j)}$ (descending) and return the indices of the top-$k$.
\end{algorithmic}
\end{algorithm}

\begin{algorithm}[ht]
\caption{End-to-End Pipeline (Train/Val/Test, Thresholding, PI, CIs; Robustness)}
\label{alg:full_pipeline}
\begin{algorithmic}[1]
\Require Dataset $\mathcal{D}$, random seed, $k$ (CDC: $k{=}10$; PIMA: $k{=}8$), models $\mathcal{M}=\{\text{RF},\text{XGB},\text{LR},\text{GB}\}$
\Ensure Test performance across feature sets; permutation importance (PI); $\lambda_U$ CIs; robustness; significance tests

\State \textbf{Split.} Stratified train/test ($80/20$). From train, hold out a validation split ($20\%$ of train) for threshold tuning.
\State \textbf{(PIMA hygiene, only for PIMA).} Treat zeros as missing in \texttt{Glucose}, \texttt{BloodPressure}, \texttt{SkinThickness}, \texttt{Insulin}, \texttt{BMI}; fit a \emph{median} imputer on train and transform both train and test (no leakage).
\State \textbf{Feature scoring on train only.} Compute $\lambda_U$ table by Alg.~\ref{alg:gumbel_feature_selection}; let \texttt{Gumbel} top-$k$ be those features. Also form baselines: \texttt{MI}, \texttt{L1/ElasticNet}, \texttt{mRMR}, \texttt{ReliefF}, and the \texttt{All} set.
\State \textbf{Bootstrap CIs for $\lambda_U$ (top-$k$).} For $B{=}1000$ resamples of train, recompute $\lambda_U$ for the selected features; report $2.5$–$97.5$ percentile CIs.
\For{each feature set $S \in \{\texttt{All},\texttt{Gumbel},\texttt{MI},\texttt{L1EN},\texttt{mRMR},\texttt{ReliefF}\}$}
  \For{each model $m \in \mathcal{M}$}
    \State \textbf{Build model.} Use class balancing (\texttt{class\_weight=`balanced'} for RF/LR/GB; \texttt{scale\_pos\_weight} for XGB). Apply isotonic calibration (5-fold) for RF/XGB/GB. Standardize inputs for LR only.
    \State \textbf{Fit on train, tune threshold on val.} Fit on train (with sample weights for GB). On val, choose the probability threshold that maximizes F1 (default; Youden optional).
    \State \textbf{Final fit \& test.} Refit on full train with $S$, apply the chosen threshold to test scores; record ROC-AUC, accuracy, precision, recall, F1; save ROC/PR curves and confusion matrix.
  \EndFor
\EndFor
\State \textbf{Pick best configuration.} By primary metric (ROC-AUC), identify the best (feature set, model).
\State \textbf{Permutation importance (PI).} 
\begin{itemize}
  \item Compute PI on the test set for the \emph{overall best} (set, model) using \texttt{permutation\_importance} with \texttt{scoring=`roc\_auc'}, $n\_repeats{=}500$.
  \item Compute PI again for the \texttt{Gumbel} top-$k$ features using the \emph{best model within Gumbel} (same scorer, $n\_repeats{=}500$).
\end{itemize}
\State \textbf{Robustness checks.} Re-evaluate the best Gumbel model under small perturbations: label flip $5\%$, feature noise $\sigma{=}0.10$, MCAR missingness $10\%$ (median impute).
\State \textbf{Significance tests.} For the best model, compare the \texttt{Gumbel} set against other sets using paired DeLong tests on ROC-AUC and McNemar’s test on correctness.
\end{algorithmic}
\end{algorithm}

\subsection*{Reproducibility}

All experiments were conducted using Google Colab Pro. The runtime environment was configured with Python 3 as the language runtime, CPU as the hardware accelerator with Colab Pro high-memory option enabled, approximately 51 GB of available RAM, and approximately 226 GB of available disk space. Because experiments were executed on Google Colab Pro (shared CPU resources), absolute wall–clock times can vary slightly across sessions. However, we consistently observe the same relative ordering of methods (the “pattern” of faster/slower selectors remains unchanged).

In the next section, we present empirical results obtained with the above pipeline and we highlight performance across the CDC and the PIMA datasets.

\section*{Results}
\label{sec:results}

In this section, we present the results of our Gumbel-$\lambda_U$ feature selection method, which was tested on two datasets: the large-scale CDC Diabetes Health Indicators survey and the clinical PIMA Indians Diabetes Database. For each dataset, we list the top features chosen by our method and the four baseline methods. Next, we compare predictive performance using key classification metrics, present statistical significance tests, and analyze feature importance for the best models. Finally, we discuss the computational complexity of each feature selection method.

\subsection*{Results for the CDC Dataset}

After running our feature‐selection methods on the CDC dataset, we identified the top 10 predictors from each approach:

\begin{itemize}
  \item \textbf{Gumbel-$\lambda_U$:} GenHlth, HighBP, DiffWalk, HighChol, BMI, HeartDiseaseorAttack, Age, PhysHlth, Stroke, CholCheck
  \item \textbf{MI:} GenHlth, HighBP, AnyHealthcare, PhysActivity, CholCheck, HighChol, Veggies, Fruits, BMI, Age
  \item \textbf{mRMR:} GenHlth, Sex, AnyHealthcare, HvyAlcoholConsump, CholCheck, Veggies, HighChol, PhysActivity, Stroke, BMI
  \item \textbf{ReliefF:} GenHlth, Age, BMI, Education, Income, HighBP, HighChol, Sex, Smoker, MentHlth
 \item \textbf{L1EN:} GenHlth, BMI, Age, HighBP, HighChol, CholCheck, HvyAlcoholConsump, Sex, Income, HeartDiseaseorAttack
\end{itemize}

\paragraph{Summary:} A clear pattern emerges from this comparison: GenHlth is unanimously selected as the top predictor by all five methods. There is also strong consensus for HighBP, HighChol, BMI, and Age, which recur across multiple lists. Notably, our Gumbel–$\lambda_U$ selector elevates DiffWalk and HeartDiseaseorAttack. Additionally, our method also identifies PhysHlth, Stroke, and CholCheck, which are features plausibly tied to severe health status and consistent with its focus on upper-tail co-occurrence with the label. The mRMR method, by contrast, prioritizes several lifestyle and healthcare-access variables (e.g., AnyHealthcare, HvyAlcoholConsump, Veggies), reflecting its criterion that penalizes redundancy among selected features.

Next, we present the classification results.

\subsubsection*{Classification Performance}

The predictive performance for each feature selection method across the four classifiers is detailed in Table~\ref{tab:cdc_performance}.

\begin{table}[ht]
\centering
\caption{Predictive performance on the CDC dataset test set for each feature selection method and classifier. The highest ROC-AUC within each feature set is bolded. ROC-AUC is reported with 95\% bootstrap confidence intervals.}
\label{tab:cdc_performance}
\begin{tabular}{l l c c c c c}
\toprule
\textbf{Set} & \textbf{Model} & \textbf{Accuracy} & \textbf{Precision} & \textbf{Recall} & \textbf{F1-Score} & \textbf{ROC-AUC (95\% CI)} \\
\midrule
\multirow{4}{*}{All} 
& GB  & 0.804 & 0.854 & 0.804 & 0.822 & 0.827 [0.823, 0.832] \\
& LR  & 0.791 & 0.852 & 0.791 & 0.813 & 0.820 [0.815, 0.825] \\
& RF  & 0.779 & 0.847 & 0.779 & 0.804 & 0.801 [0.796, 0.806] \\
& XGB & 0.809 & 0.853 & 0.809 & 0.826 & 0.827 [0.823, 0.832] \\
\midrule
\multirow{4}{*}{Gumbel} 
& GB  & 0.813 & 0.850 & 0.813 & 0.827 & \textbf{0.823} [0.819, 0.828] \\
& LR  & 0.797 & 0.849 & 0.797 & 0.817 & 0.816 [0.812, 0.821] \\
& RF  & 0.732 & 0.826 & 0.732 & 0.766 & 0.741 [0.735, 0.746] \\
& XGB & 0.810 & 0.851 & 0.810 & 0.826 & 0.822 [0.817, 0.827] \\
\midrule
\multirow{4}{*}{MI} 
& GB  & 0.821 & 0.848 & 0.821 & 0.832 & 0.821 [0.817, 0.826] \\
& LR  & 0.793 & 0.849 & 0.793 & 0.814 & 0.815 [0.810, 0.820] \\
& RF  & 0.733 & 0.833 & 0.733 & 0.768 & 0.755 [0.748, 0.760] \\
& XGB & 0.814 & 0.849 & 0.814 & 0.828 & 0.820 [0.815, 0.825] \\
\midrule
\multirow{4}{*}{mRMR} 
& GB  & 0.781 & 0.843 & 0.781 & 0.804 & 0.794 [0.788, 0.799] \\
& LR  & 0.763 & 0.843 & 0.763 & 0.791 & 0.789 [0.784, 0.795] \\
& RF  & 0.745 & 0.839 & 0.745 & 0.777 & 0.763 [0.757, 0.769] \\
& XGB & 0.781 & 0.842 & 0.781 & 0.804 & 0.792 [0.787, 0.797] \\
\midrule
\multirow{4}{*}{ReliefF} 
& GB  & 0.805 & 0.851 & 0.805 & 0.822 & 0.822 [0.818, 0.827] \\
& LR  & 0.791 & 0.850 & 0.791 & 0.813 & 0.815 [0.810, 0.820] \\
& RF  & 0.748 & 0.835 & 0.748 & 0.779 & 0.766 [0.760, 0.771] \\
& XGB & 0.814 & 0.849 & 0.814 & 0.828 & 0.821 [0.816, 0.826] \\
\midrule
\multirow{4}{*}{L1EN} 
& GB  & 0.801 & 0.855 & 0.801 & 0.820 & 0.826 [0.822, 0.831] \\
& LR  & 0.804 & 0.850 & 0.804 & 0.822 & 0.819 [0.815, 0.824] \\
& RF  & 0.733 & 0.833 & 0.733 & 0.768 & 0.753 [0.747, 0.759] \\
& XGB & 0.809 & 0.852 & 0.809 & 0.825 & 0.825 [0.820, 0.830] \\
\bottomrule
\end{tabular}
\end{table}

We report ROC-AUC with 95\% bootstrap confidence intervals (\(B=1000\) resamples) to quantify uncertainty under resampling and to complement paired significance tests.

Using all 21 features gives the best overall performance, which is common in datasets with a moderate number of predictors. Moreover, the \texttt{All}, Gumbel-$\lambda_U$, and L1EN configurations yield closely matched ROC-AUC values, and their 95\% bootstrap confidence intervals frequently overlap, indicating that the observed differences are small relative to resampling variability. Our main focus, however, is to assess how well the selection methods perform. The results in Table~\ref{tab:cdc_performance} show a clear ranking: our Gumbel-$\lambda_U$ method and the L1/Elastic-Net (L1EN) baseline consistently perform among the top methods across classifiers, with strong and nearly identical results. For example, with Gradient Boosting (GB), Gumbel-$\lambda_U$ attains a ROC-AUC of 0.823, which is close to L1EN’s 0.826. Within the Gumbel-$\lambda_U$ feature set, GB achieves the highest ROC-AUC (0.823), making it the best-performing classifier for our proposed selected features on the CDC dataset.

The Gumbel-$\lambda_U$ feature set shows a clear advantage over several other filter methods. Its ROC-AUC of 0.823 is numerically higher than the scores for Mutual Information (MI) at 0.821 and mRMR at 0.794. As the formal statistical tests in the next section will confirm, our method is not only competitive but statistically higher ROC-AUC than MI and mRMR on this split, suggesting that the upper-tail criterion can be a useful screening signal in this setting.

\begin{figure}[ht]
    \centering
    \includegraphics[width=0.6\linewidth]{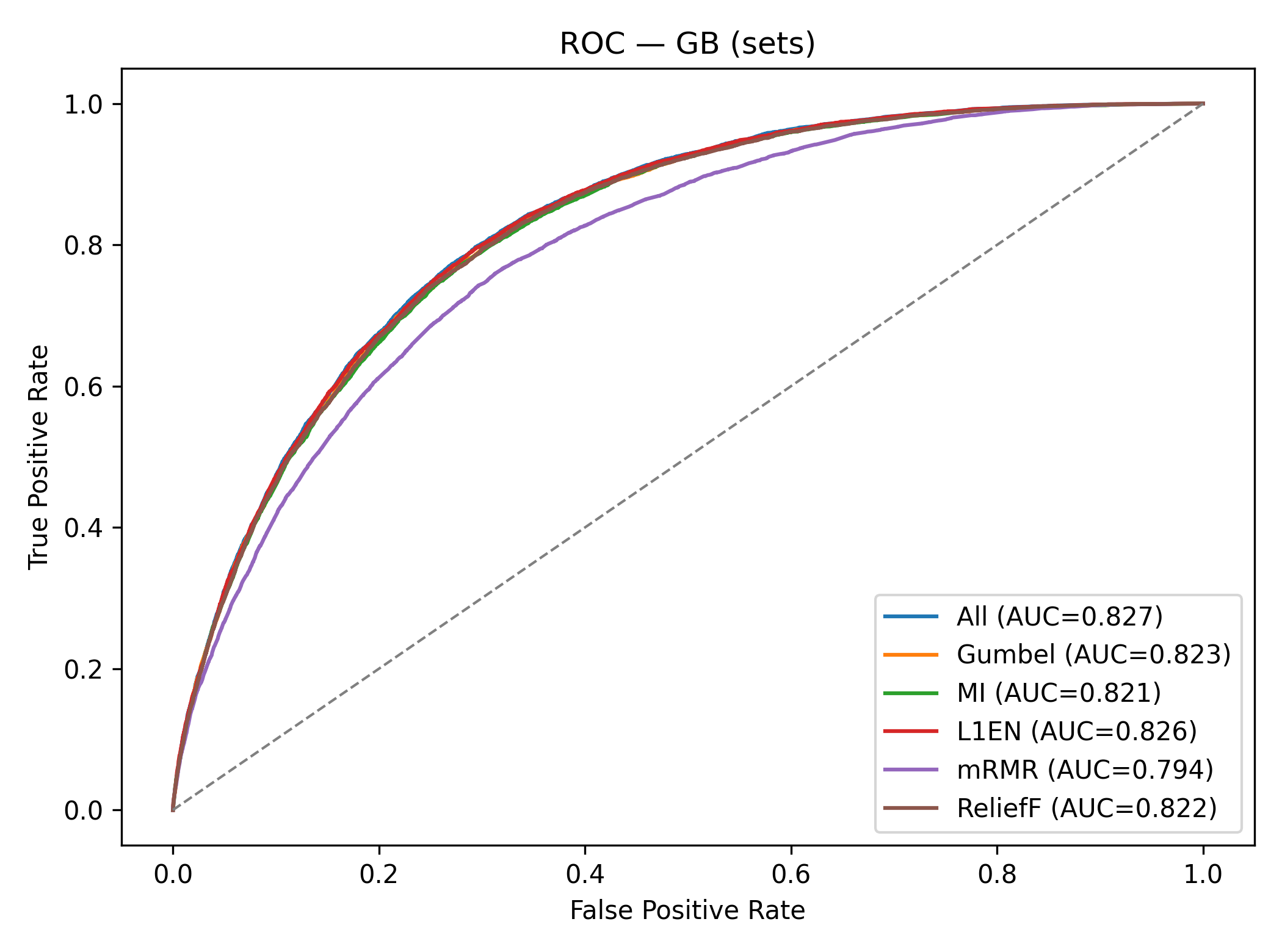}
    \caption{ROC curves for GB across feature sets on CDC (AUCs in legend); Gumbel–$\lambda_U$ closely tracks the top performers.}
    \label{fig:cdc_roc_gb}
\end{figure}

To provide a visual comparison of these results, Figure~\ref{fig:cdc_roc_gb} displays the superimposed ROC curves for the top-performing Gradient Boosting model across each feature set. The plot clearly shows the curves for our Gumbel-$\lambda_U$ method, MI, L1EN, ReliefF and the All Features baseline clustered tightly together, visually confirming their very similar performance.. In contrast, the curve for mRMR falls visibly below this top group, consistent with their lower AUC score reported in Table~\ref{tab:cdc_performance}. This visualization reinforces the conclusion that our proposed method effectively identifies a feature set that is highly competitive with the best-performing baselines.

To determine if the small differences between our method and the top baselines are statistically significant, we now present the results of paired statistical tests.

\subsubsection*{Statistical Tests}
\begin{figure}[ht]
    \centering
    \includegraphics[width=0.6\linewidth]{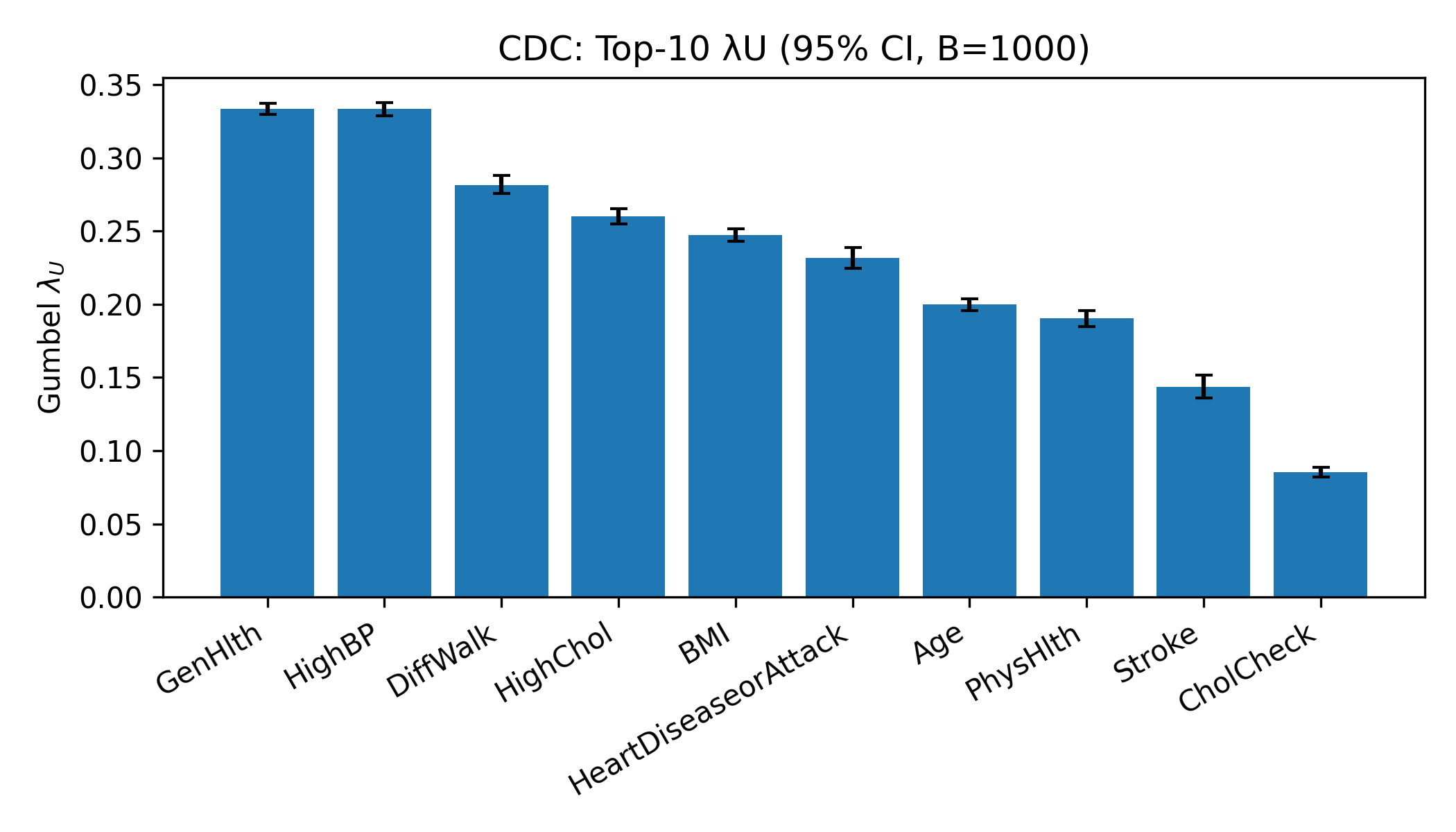}
    \caption{CDC: top-10 Gumbel $\lambda_U$ with 95\% bootstrap confidence intervals ($B{=}1000$).}
    \label{fig:lambdaU_CIs}
\end{figure}

To validate the Gumbel-$\lambda_U$ scores and assess the statistical significance of our model's performance, we conducted bootstrap analysis and paired statistical tests. First, we generated 95\% bootstrap confidence intervals for the $\lambda_U$ scores of the top-10 selected features (Figure~\ref{fig:lambdaU_CIs}). The intervals for all top-ranked predictors lie strictly above zero, indicating that our method is capturing a genuine upper-tail dependence signal rather than noise.

Now we use a covariance-aware DeLong’s test to compare ROC-AUCs in a paired manner. The model trained on all 21 features achieved the highest performance (ROC-AUC = 0.827). As expected given the modest full set (21 variables) and our emphasis on parsimony, we focus comparisons among selectors that return exactly 10 features (k = 10).

From Table~\ref{tab:cdc_stats_tests}, we see that our Gumbel-\(\lambda_U\) filter reduced the feature space by \(\approx\)52\% (21\(\to\)10) and yielded a competitive ROC-AUC of 0.823. The difference vs. All is statistically significant, but the key comparison is among 10-feature selectors: Gumbel-\(\lambda_U\) is better than MI \(\left(p=1.47\times10^{-5}\right)\) and mRMR \(\left(p<0.001\right)\), and statistically indistinguishable from ReliefF \(\left(p=0.154\right)\). 
L1EN shows a statistically detectable but numerically small AUC difference relative to Gumbel ($p = 2.69 \times 10^{-9}$; $\Delta \text{AUC} \approx 0.003$), consistent with its embedded, model-coupled nature.

Finally, McNemar’s test on paired error profiles is highly significant for Gumbel vs.\ each baseline (Table~\ref{tab:cdc_stats_tests}), indicating the models make different types of mistakes even when AUCs are similar.

\textit{Note.} Extremely small p-values are reported as $p<0.001$ instead of 0.

\begin{table}[H]
\centering
\caption{Paired tests on the CDC test set comparing the Gumbel-selected feature set against alternatives (Gradient Boosting). DeLong compares ROC-AUC; McNemar compares error profiles.}
\label{tab:cdc_stats_tests}
\begin{tabular}{lccc}
\toprule
\textbf{Comparison} & \textbf{DeLong $p$} & \textbf{McNemar $\chi^2$} & \textbf{McNemar $p$} \\
\midrule
Gumbel vs.\ All     & $<0.001$ & 94.86  & $2.05\times10^{-22}$ \\
Gumbel vs.\ MI      & $1.47\times10^{-5}$ & 135.96 & $2.03\times10^{-31}$ \\
Gumbel vs.\ mRMR    & $<0.001$ & 496.05 & $6.89\times10^{-110}$ \\
Gumbel vs.\ ReliefF & $0.154$     & 70.30  & $5.10\times10^{-17}$ \\
Gumbel vs.\ L1EN   & $2.69\times10^{-9}$ & 158.02 & $3.07\times10^{-36}$ \\
\bottomrule
\end{tabular}
\end{table}

Next, we look at the permutation importance within the Gumbel set.

\subsubsection*{Permutation Importance Within the Gumbel Set (CDC)}
\begin{figure}[ht]
    \centering
    \includegraphics[width=0.6\linewidth]{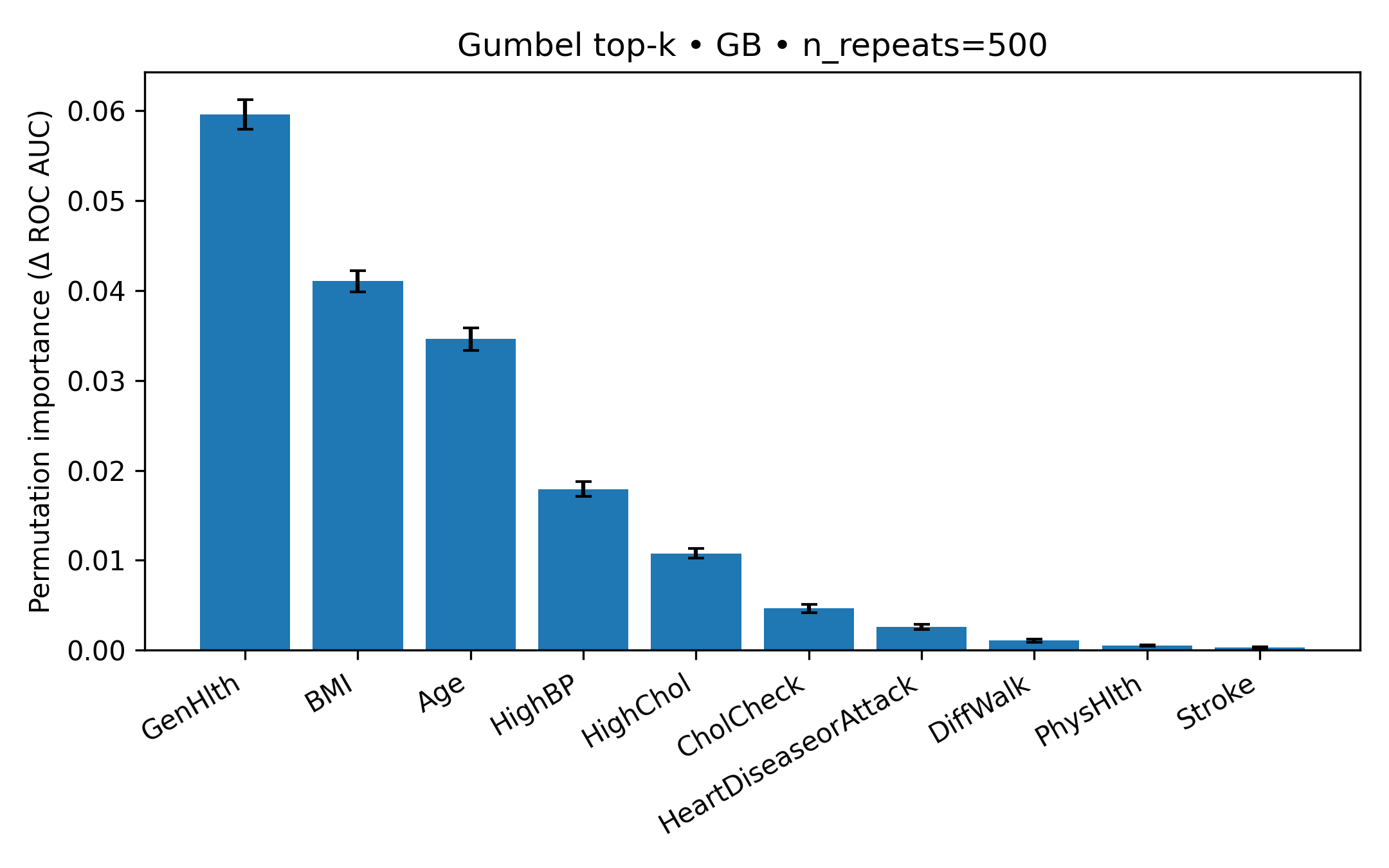}
    \caption{Permutation importance (mean $\Delta$ROC-AUC over 500 permutations) for Gradient Boosting on the Gumbel top-10 features (CDC test set). Larger values indicate greater importance.}
    \label{fig:cdc_perm_importance}
\end{figure}

To understand which variables our method uses, we computed permutation importance for our best CDC setup: Gradient Boosting trained on the Gumbel top-10 features (Figure~\ref{fig:cdc_perm_importance}). On the held-out CDC test set, we permuted one feature at a time (500 repetitions) and measured the mean drop in ROC-AUC. Larger drops mean the model relies more on that feature.

The results, visualized in Figure~\ref{fig:cdc_perm_importance} and detailed in Table~\ref{tab:cdc_perm_importance}, reveal a clear hierarchy of feature importance. GenHlth is the most influential predictor by a wide margin. BMI and Age follow with strong contributions, and HighBP and HighChol provide additional signal. Smaller but non-zero effects are observed for CholCheck and HeartDiseaseorAttack, while DiffWalk, PhysHlth, and Stroke contribute modestly. All ten Gumbel-selected features have positive mean importance, confirming that each contributes meaningfully to the model’s performance.

\begin{table}[ht]
\centering
\caption{Permutation importance of the top-10 features selected by the Gumbel-$\lambda_U$ method for the CDC dataset. Importance is measured as the mean decrease in ROC-AUC over 500 permutations on the test set, using the trained Gradient Boosting model.}
\label{tab:cdc_perm_importance}
\begin{tabular}{lcc}
\toprule
\textbf{Feature} & \textbf{Mean Importance ($\Delta$ ROC-AUC)} & \textbf{Std. Deviation} \\
\midrule
GenHlth               & 0.059574 & 0.001662 \\
BMI                   & 0.041051 & 0.001183 \\
Age                   & 0.034596 & 0.001264 \\
HighBP                & 0.017929 & 0.000814 \\
HighChol              & 0.010755 & 0.000545 \\
CholCheck             & 0.004652 & 0.000475 \\
HeartDiseaseorAttack  & 0.002608 & 0.000268 \\
DiffWalk              & 0.001073 & 0.000197 \\
PhysHlth              & 0.000501 & 0.000082 \\
Stroke                & 0.000303 & 0.000108 \\
\bottomrule
\end{tabular}
\end{table}

After this, we evaluated the stability of our Gumbel-selected feature set under several common data perturbations.

\subsubsection*{Robustness Checks}
We stress–tested the Gradient Boosting (GB) classifier trained on the Gumbel top-10 features under three perturbations (Table~\ref{tab:cdc_robust}). Relative to the baseline ROC-AUC of 0.823, performance remained within $\approx 0.001$ under label–flip 5\% (0.8223) and feature noise $\sigma{=}0.10$ (0.8227), and declined by $\approx 0.011$ under MCAR 10\% (0.8127), confirming robustness under light noise.

\begin{table}[ht]
\centering
\caption{CDC robustness: GB on Gumbel top-10.}
\label{tab:cdc_robust}
\begin{tabular}{lcc}
\toprule
\textbf{Type} & \textbf{Level} & \textbf{ROC-AUC} \\
\midrule
Baseline      & 0.00 & 0.82321 \\
Label flip    & 0.05 & 0.82230 \\
Feature noise ($\sigma$) & 0.10 & 0.82270 \\
MCAR          & 0.10 & 0.81266 \\
\bottomrule
\end{tabular}
\end{table}

Having confirmed that our method's performance is stable under noisy conditions, we now check the computational complexity of the feature selectors.

\subsubsection*{Runtime of Feature Selectors}
Asymptotically the selector is $O(d\,n\log n)$ as detailed in Preliminaries; wall-clock times below reflect that efficiency on CDC.
 We recorded wall-clock time for each selector on the CDC training split (Table~\ref{tab:cdc_fs_time}). Gumbel–$\lambda_U$ is fastest (0.332 s), about $9\times$ faster than L1EN and $\approx 61\times$ faster than MI and mRMR; ReliefF is orders of magnitude slower ($\sim 2.8\times 10^3\times$).

\begin{table}[ht]
\centering
\caption{Feature–selection wall-clock time (seconds) on the CDC dataset.}
\label{tab:cdc_fs_time}
\begin{tabular}{lc}
\toprule
\textbf{Selector} & \textbf{Time (s)} \\
\midrule
Gumbel–$\lambda_U$ & 0.332 \\
L1EN    & 3.102 \\
MI & 20.205 \\
mRMR               & 20.515 \\
ReliefF            & 925.591 \\
\bottomrule
\end{tabular}
\end{table}

\paragraph{Comment on CDC feature type and model behavior.} The CDC dataset contains a large number of binary/discrete predictors. In such settings, a linear model can perform competitively because much of the apparent nonlinearity is effectively represented through indicator coding, which helps explain why logistic regression can match or exceed tree-based models in our benchmarks. As a result, while the CDC experiment provides strong evidence that the proposed upper-tail screening is effective and efficient, further evaluation on datasets with richer continuous predictors and stronger interaction-driven nonlinearities is a natural direction for future work.

Next, we move to the PIMA dataset.

\subsection*{Results for the PIMA Dataset}

PIMA serves as a complementary clinical benchmark where no dimensionality reduction is possible (eight total predictors). Consequently, all methods use the same variables; differences reflect sampling and calibration randomness (column order has no effect for LR/RF/GB/XGB). Under this setup, the Gumbel ranking paired with Random Forest achieved the \emph{numerically highest} ROC-AUC (0.867). Formal paired tests are reported in the Statistical Tests subsection.

After running our feature‐selection methods on the PIMA dataset, we identified the top 8 predictors (which are same as all the features) from each approach:

\begin{itemize}
  \item \textbf{Gumbel-$\lambda_U$:} Glucose, BMI, Age, Insulin, SkinThickness, Pregnancies, BloodPressure, DiabetesPedigreeFunction
  \item \textbf{MI:}Glucose, BMI, Age, BloodPressure, Insulin, Pregnancies, SkinThickness, DiabetesPedigreeFunction
 \item \textbf{mRMR:}Glucose, Pregnancies, BMI, DiabetesPedigreeFunction, Insulin, BloodPressure, Age, SkinThickness
  \item \textbf{ReliefF:}Glucose, BMI, SkinThickness, BloodPressure, Age, Pregnancies, DiabetesPedigreeFunction, Insulin
 \item \textbf{L1EN:}Glucose, BMI, Pregnancies, DiabetesPedigreeFunction, Insulin, BloodPressure, Age, SkinThickness

\end{itemize}

\paragraph{Summary:}A clear pattern emerges for PIMA: Glucose stands out as the top predictor in all five methods. There is also strong agreement on the importance of BMI and Age, which consistently rank near the top, while BloodPressure and Pregnancies appear frequently as well. Our Gumbel-$\lambda_U$ selector ranks Insulin and SkinThickness higher than some other methods, which matches its focus on upper-tail co-occurrence with the label, meaning it highlights extreme metabolic readings that align with positive outcomes. In contrast, mRMR produces a slightly different order by bringing DiabetesPedigreeFunction forward, reflecting its approach to reducing redundancy. ReliefF gives more weight to SkinThickness because of its ability to separate classes locally. Since the PIMA dataset has eight predictors, all methods keep the full set, so the main difference is in the ranking. The Gumbel method highlights clinically relevant factors like glycemic control and adiposity as key drivers of diabetes risk.

After this, we present the classification results.

\subsubsection*{Classification Performance}
PIMA has only 8 predictors, so feature selection here is more about ranking than reducing dimensions. We include it as a clinical benchmark to check that our tail-dependence criterion works as expected on a small, well-studied dataset, to compare with standard selectors using the same training and thresholds, and to show that our method performs as well as or better than others without needing many features. This approach complements the CDC results, where dimensionality reduction plays a larger role.

The predictive performance for each feature ranking on the PIMA dataset is presented in Table~\ref{tab:pima_performance}. A unique characteristic of this benchmark is that it contains only eight predictors, so every method ultimately ranks the full feature set. Because every model trains on exactly the same eight variables, AUCs are expected to be very close. The small differences observed reflect sampling/calibration randomness rather than meaningful effects, which, as later confirmed by our DeLong tests (all p > 0.05), are not significant.

\begin{table}[ht]
\centering
\caption{Predictive performance on the PIMA dataset test set for each feature ranking and classifier. We report ROC-AUC with 95\% bootstrap confidence intervals. The highest ROC-AUC within each feature set is bolded.}
\label{tab:pima_performance}
\begin{tabular}{l l c c c c c}
\toprule
\textbf{Set} & \textbf{Model} & \textbf{Accuracy} & \textbf{Precision} & \textbf{Recall} & \textbf{F1} & \textbf{ROC-AUC (95\% CI)} \\
\midrule
\multirow{4}{*}{All}
& GB  & 0.773 & 0.806 & 0.773 & 0.778 & 0.860 [0.791, 0.916] \\
& LR  & 0.773 & 0.768 & 0.773 & 0.766 & 0.862 [0.796, 0.913] \\
& RF  & 0.766 & 0.792 & 0.766 & 0.771 & 0.861 [0.798, 0.918] \\
& XGB & 0.779 & 0.786 & 0.779 & 0.782 & 0.840 [0.767, 0.899] \\
\midrule
\multirow{4}{*}{Gumbel}
& GB  & 0.682 & 0.790 & 0.682 & 0.685 & 0.858 [0.789, 0.914] \\
& LR  & 0.773 & 0.768 & 0.773 & 0.766 & 0.862 [0.796, 0.913] \\
& RF  & 0.753 & 0.801 & 0.753 & 0.759 & \textbf{0.867} [0.804, 0.920] \\
& XGB & 0.760 & 0.804 & 0.760 & 0.765 & 0.824 [0.748, 0.885] \\
\midrule
\multirow{4}{*}{MI}
& GB  & 0.773 & 0.806 & 0.773 & 0.778 & 0.859 [0.789, 0.914] \\
& LR  & 0.773 & 0.768 & 0.773 & 0.766 & 0.862 [0.796, 0.913] \\
& RF  & 0.812 & 0.817 & 0.812 & 0.813 & 0.859 [0.794, 0.912] \\
& XGB & 0.747 & 0.776 & 0.747 & 0.752 & 0.842 [0.769, 0.902] \\
\midrule
\multirow{4}{*}{mRMR}
& GB  & 0.760 & 0.799 & 0.760 & 0.765 & 0.863 [0.795, 0.919] \\
& LR  & 0.773 & 0.768 & 0.773 & 0.766 & 0.862 [0.796, 0.913] \\
& RF  & 0.760 & 0.799 & 0.760 & 0.765 & 0.863 [0.796, 0.917] \\
& XGB & 0.753 & 0.784 & 0.753 & 0.759 & 0.839 [0.768, 0.898] \\
\midrule
\multirow{4}{*}{ReliefF}
& GB  & 0.779 & 0.815 & 0.779 & 0.784 & 0.866 [0.799, 0.920] \\
& LR  & 0.773 & 0.768 & 0.773 & 0.766 & 0.862 [0.796, 0.913] \\
& RF  & 0.812 & 0.813 & 0.812 & 0.812 & 0.860 [0.795, 0.915] \\
& XGB & 0.799 & 0.804 & 0.799 & 0.801 & 0.835 [0.765, 0.896] \\
\midrule
\multirow{4}{*}{L1EN}
& GB  & 0.760 & 0.817 & 0.760 & 0.765 & 0.864 [0.799, 0.918] \\
& LR  & 0.773 & 0.768 & 0.773 & 0.766 & 0.862 [0.796, 0.913] \\
& RF  & 0.792 & 0.809 & 0.792 & 0.796 & 0.861 [0.800, 0.913] \\
& XGB & 0.753 & 0.775 & 0.753 & 0.758 & 0.833 [0.763, 0.894] \\
\bottomrule
\end{tabular}
\end{table}

We report ROC-AUC with 95\% bootstrap confidence intervals (\(B=1000\) resamples) to quantify uncertainty under resampling and to complement paired significance tests.

On PIMA (8 predictors), dimensionality reduction is not the goal, so this experiment primarily serves as a ranking sanity check in a low-dimensional clinical setting. In Table~\ref{tab:pima_performance}, the best-performing configurations across \texttt{All}, Gumbel-$\lambda_U$, and strong baselines such as ReliefF and L1EN are very close, and the 95\% ROC-AUC bootstrap confidence intervals overlap substantially, consistent with the expectation that differences largely reflect sampling variability (and modest model-fitting/calibration effects) rather than screening gains. Nevertheless, Gumbel-$\lambda_U$ paired with Random Forest achieves the numerically highest ROC-AUC (0.867; 95\% CI [0.804, 0.920]), indicating that the tail-aware ranking remains competitive and clinically coherent even when dimensionality reduction is not possible.

The wider ROC-AUC confidence intervals on the Pima dataset are expected due to its smaller sample size, which increases sampling variability; conversely, the larger CDC cohort yields tighter intervals.

\begin{figure}[ht]
    \centering
    \includegraphics[width=0.6\linewidth]{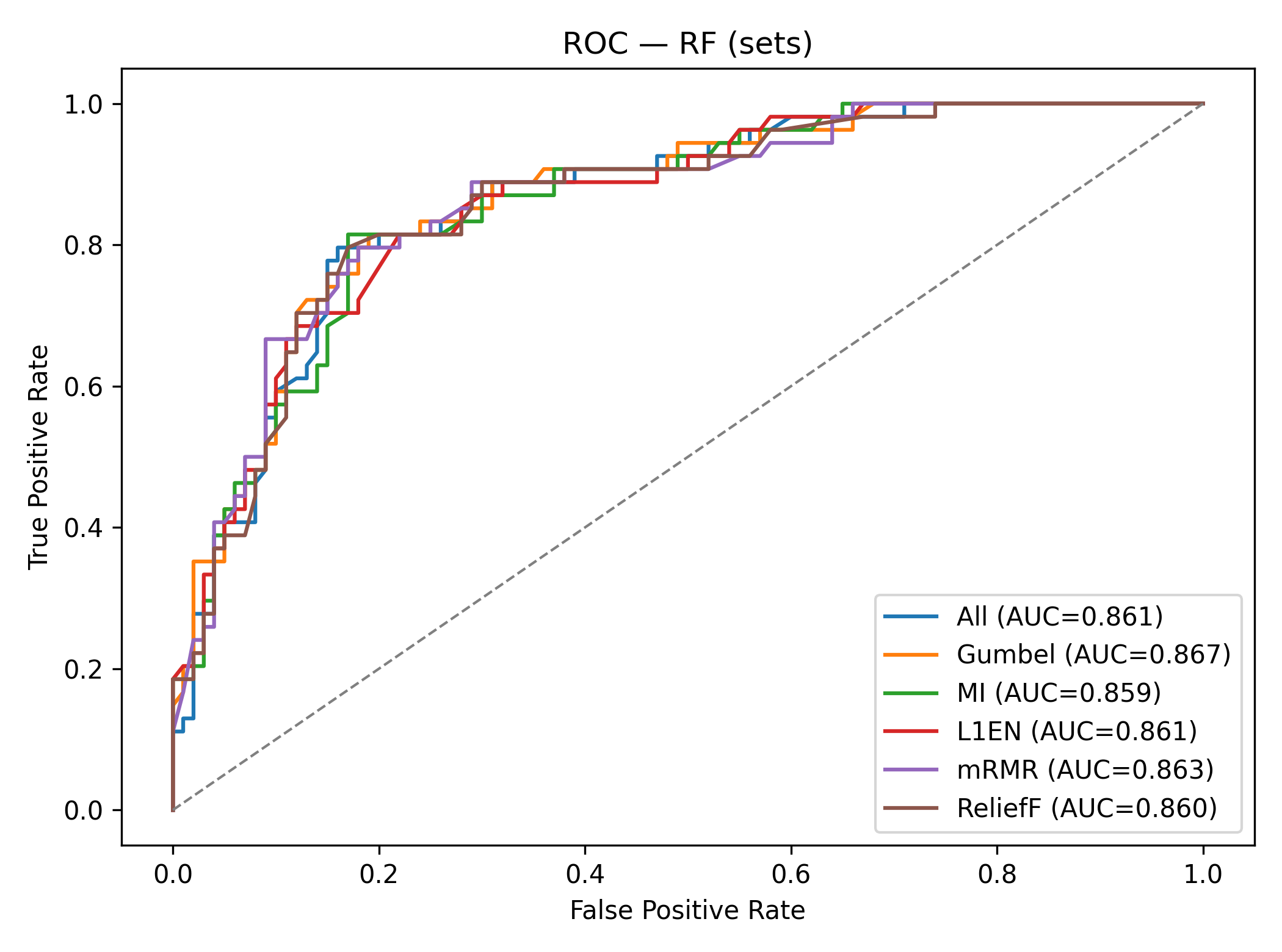}
    \caption{Superimposed ROC curves for Random Forest on PIMA across feature sets. Gumbel-$\lambda_U$ yields the top AUC (0.867); all others are close, underscoring minimal sensitivity to feature set size on this 8-variable benchmark.}
    \label{fig:pima_roc_rf}
\end{figure}

To visualize the PIMA results, Figure~\ref{fig:pima_roc_rf} shows superimposed ROC curves for the Random Forest model across all feature sets. The curves are tightly clustered, indicating very similar discrimination. The Gumbel-$\lambda_U$ set attains the numerically highest AUC (0.867), with mRMR, All, L1EN, and ReliefF close behind and MI slightly lower.

A key finding is that even with a small number of features where feature selection is not needed to manage model complexity, our upper-tail dependence ranking still performs at a high level. It highlights clinically relevant factors such as Glucose, BMI, and Age, supporting CDC findings that focusing on joint extremes can be informative.

To determine if the small differences between our method and the top baselines are statistically significant, we now present the results of paired statistical tests.

\emph{Note:} PIMA has 8 predictors; all methods use the same variables. Reported differences therefore arise from ranked ordering and model refitting/calibration rather than feature removal.

\subsubsection*{Statistical Tests}
\begin{figure}[ht]
    \centering
    \includegraphics[width=0.6\linewidth]{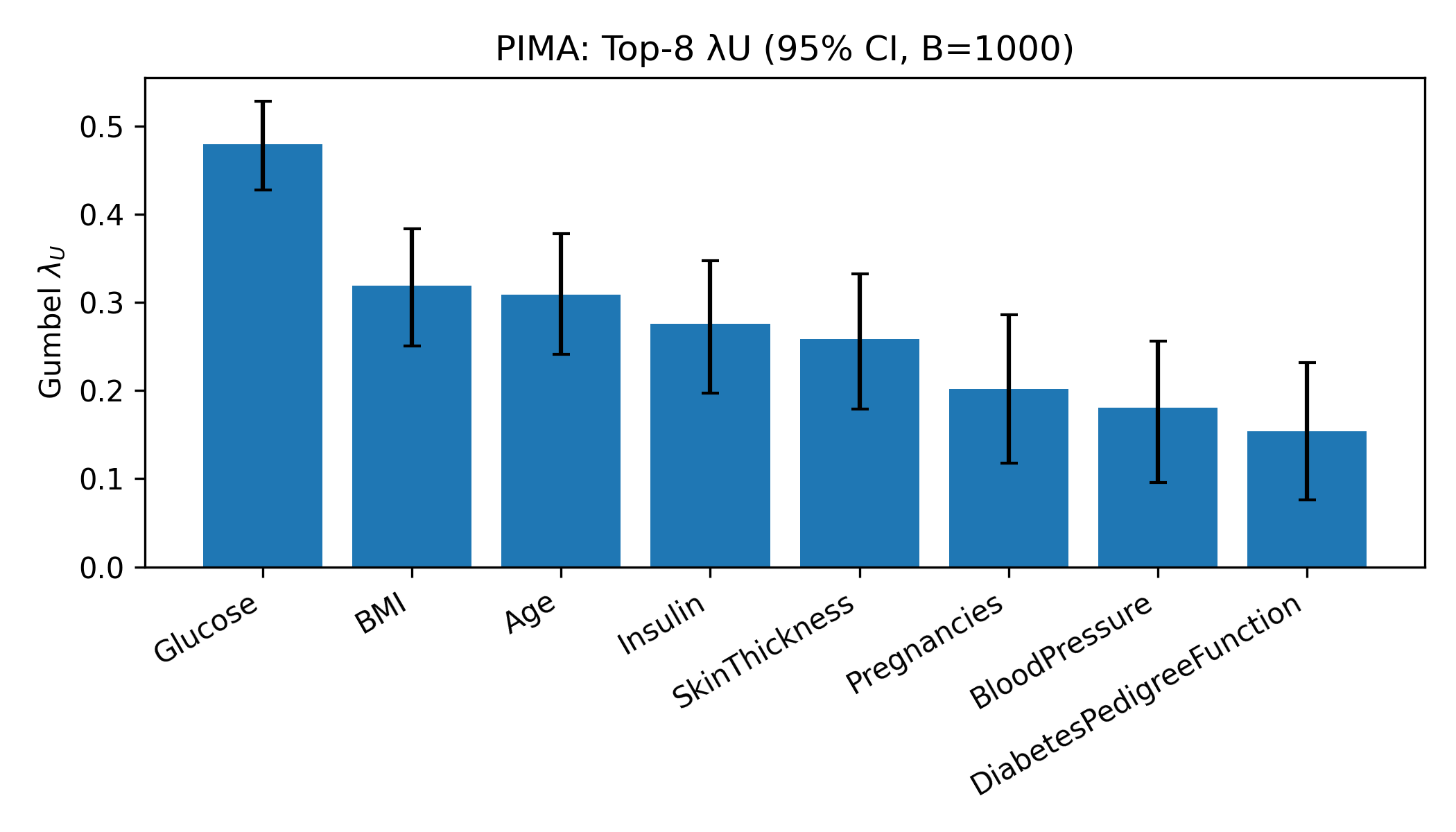}
    \caption{PIMA: Top-8 $\lambda_U$ with 95\% bootstrap CIs ($B{=}1000$).}
    \label{fig:pima_lambdaU_CIs}
\end{figure}

To validate the Gumbel-$\lambda_U$ feature rankings and assess the statistical significance of our model's performance on the PIMA dataset, we again conducted bootstrap analysis and paired statistical tests. First, we generated 95\% bootstrap confidence intervals for the $\lambda_U$ scores of the top-ranked features (see Figure~\ref{fig:pima_lambdaU_CIs}). As with the CDC data, the confidence intervals for the clinically important predictors like Glucose, BMI, and Age are strictly greater than zero, confirming that our method is capturing a genuine and statistically significant upper-tail dependence signal.

Next, we used a covariance-aware DeLong's test to compare the ROC-AUC of the best-performing classifier (Random Forest) trained on the Gumbel-$\lambda_U$ features against each baseline. As shown in Table~\ref{tab:pima_stats_tests}, all pairwise comparisons yielded $p>0.05$. Therefore, we fail to reject the null in every case and conclude there is no statistically significant difference in AUC. This supports that, in this low-dimensional clinical setting, the Gumbel-$\lambda_U$ ranking achieves performance statistically indistinguishable from strong baselines.

Interestingly, as seen from Table~\ref{tab:pima_stats_tests}, in contrast to the CDC results, McNemar's test also returned non-significant p-values (p > 0.05) for all comparisons. This indicates that on this smaller, more homogeneous clinical dataset, the different feature rankings not only produce similar overall accuracy but also share a similar pattern of classification errors.

\begin{table}[ht]
\centering
\caption{Paired statistical test results comparing the Gumbel-ranked feature set against all others on the PIMA test set, using the Random Forest classifier.}
\label{tab:pima_stats_tests}
\begin{tabular}{lccc}
\toprule
\textbf{Comparison} & \textbf{DeLong p-value} & \textbf{McNemar $\chi^2$} & \textbf{McNemar p-value} \\
\midrule
Gumbel vs. All Features & 0.253 & 0.13 & 0.724 \\
Gumbel vs. MI         &  0.095 & 3.37 & 0.066 \\
Gumbel vs. mRMR       &  0.504 & 0.00 & 1.000 \\
Gumbel vs. ReliefF    &  0.101 & 2.78 & 0.095 \\
Gumbel vs. L1EN      &  0.255 & 2.08 & 0.149 \\

\bottomrule
\end{tabular}
\end{table}

After this, we look at the permutation importance within the Gumbel set.

\subsubsection*{Permutation Importance Within the Gumbel Set (PIMA)}

\begin{figure}[ht]
    \centering
    \includegraphics[width=0.5\linewidth]{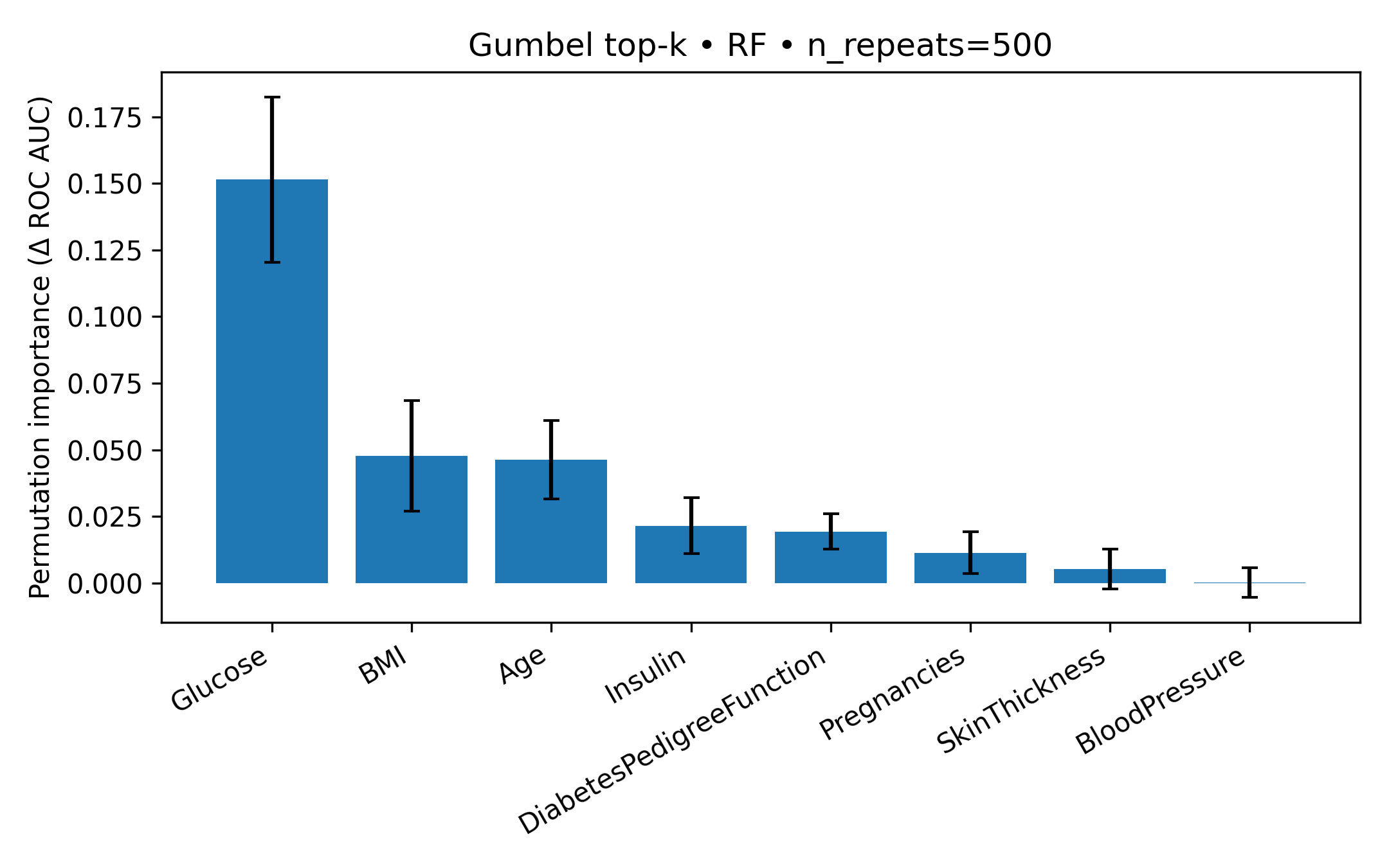}
    \caption{Permutation importance (mean $\Delta$ ROC-AUC; $\pm$1 SD over 500 permutations) for Random Forest on the PIMA Gumbel--$\lambda_U$ feature set.}
    \label{fig:pima_perm_importance}
\end{figure}

We computed permutation importance on the PIMA test set using the best configuration, Random Forest trained on the eight Gumbel--$\lambda_U$ features, by permuting each feature 500 times and measuring the mean drop in ROC-AUC (error bars show $\pm$1 SD across permutations). As shown in Figure~\ref{fig:pima_perm_importance} and detailed in Table~\ref{tab:pima_perm_importance}, Glucose dominates: its importance is a little over three times that of the next features (BMI and Age). Insulin, DiabetesPedigreeFunction, and Pregnancies provide smaller but positive contributions, while SkinThickness is marginal and BloodPressure is effectively zero. These results align with clinical expectations and confirm that the Gumbel-$\lambda_U$ ranking concentrates signal on the most informative predictors.

\begin{table}[ht]
\centering
\caption{Permutation importance of the Gumbel-ranked features for the PIMA dataset. Importance is measured as the mean decrease in ROC-AUC over 500 permutations on the test set, using the trained Random Forest model.}
\label{tab:pima_perm_importance}
\begin{tabular}{lcc}
\toprule
\textbf{Feature} & \textbf{Mean Importance ($\Delta$ ROC-AUC)} & \textbf{Std. Deviation} \\
\midrule
Glucose & 0.1514 & 0.0310 \\
BMI & 0.0477 & 0.0208 \\
Age & 0.0463 & 0.0147 \\
Insulin & 0.0216 & 0.0105 \\
DiabetesPedigreeFunction & 0.0193 & 0.0066 \\
Pregnancies & 0.0114 & 0.0078 \\
SkinThickness & 0.0054 & 0.0075 \\
BloodPressure & 0.0002 & 0.0055 \\
\bottomrule
\end{tabular}
\end{table}

Next, we evaluated the stability of our Gumbel-selected feature set under several common data perturbations.

\subsubsection*{Robustness Checks}

We stress-tested the Random Forest trained on the Gumbel top-8 features under three perturbations (Table~\ref{tab:pima_robust}). Relative to the baseline ROC-AUC of 0.8666, performance declined by about 0.015 with label-flip 5\% (0.8519), was essentially unchanged (slightly higher) with feature noise $\sigma = 0.10$ (0.8676), and declined by about 0.007 under MCAR 10\% (0.8597). These small ($\leq 0.015$ AUC) shifts indicate the model remains robust to light label noise, moderate feature perturbations, and modest missingness.

\begin{table}[ht]
\centering
\caption{PIMA robustness: RF on Gumbel-ranked features.}
\label{tab:pima_robust}
\begin{tabular}{lcc}
\toprule
\textbf{Type} & \textbf{Level} & \textbf{ROC-AUC} \\
\midrule
Baseline      & 0.00   & 0.8666 \\
Label flip    & 0.05   & 0.8519 \\
Feature noise ($\sigma$) & 0.10   & 0.8676 \\
MCAR          & 0.10   & 0.8597 \\
\bottomrule
\end{tabular}
\end{table}

Having confirmed that our method’s performance is stable under noisy conditions, we now check the computational complexity of the feature selectors.

\subsubsection*{Runtime of Feature Selectors}
As above, the selector is $O(d\,n\log n)$; on the small PIMA split, all filter methods are within the same order of magnitude. We recorded the wall-clock time for each selector on the PIMA training split (Table~\ref{tab:pima_fs_time}). On this smaller dataset, L1/Elastic-Net is the fastest at 0.0049 s, while ReliefF remains the slowest at 0.4711 s. Crucially, our Gumbel-$\lambda_U$ method is comparably fast to other standard filters like MI and mRMR (all in the $\sim$0.02--0.03 s range), demonstrating competitive predictive performance on this dataset without any significant computational penalty.

\begin{table}[ht]
\centering
\caption{Feature–selection wall-clock time (seconds) on the PIMA dataset.}
\label{tab:pima_fs_time}
\begin{tabular}{lc}
\toprule
\textbf{Selector} & \textbf{Time (s)} \\
\midrule
L1EN & 0.0049 \\
MI & 0.0221 \\
mRMR & 0.0283 \\
Gumbel–$\lambda_U$ & 0.0291 \\
ReliefF & 0.4711 \\
\bottomrule
\end{tabular}
\end{table}

This successful validation on the PIMA clinical benchmark concludes our quantitative analysis. Next, we discuss the medical and public health implications of our findings for both datasets.

\noindent\textbf{Limitation.}
PIMA’s low dimensionality limits conclusions about feature reduction; future work will assess richer multi-center EHR cohorts to evaluate tail-dependent selection under higher dimensionality and domain shift.

\paragraph{Two regimes, complementary evidence.}
On CDC (21 features), upper-tail filtering reduces the feature set by $\sim$52\%. This introduces a minor but statistically significant performance trade-off against using the full feature set, but our method's key advantage is highlighted by its statistically higher performance compared to other filter baselines, demonstrating its efficiency and ranking power. On PIMA (8 features), where selection is not applicable, our criterion yields clinically coherent rankings and maintains discrimination (no significant AUC differences), i.e., it does no harm in low-dimensional clinical settings. Together, these results support the method’s robustness, interpretability, and practicality across dataset types.

\subsection*{Medical and Public Health Implications}

Our feature selector ranks variables using the Gumbel copula upper-tail dependence $\lambda_U(X,Y)$. This measures how likely it is that a predictor $X$ and the diabetes label $Y$ are both extreme at the same time. A high $\lambda_U$ highlights variables that appear with diabetes in the highest-risk groups, which is exactly where screening and intervention are most important.

\paragraph{Clarification (clinical practice vs study contribution).}
The points below (for both CDC and PIMA) are not presented as new clinical guidelines. We emphasize that our study does not propose new interventions; rather, it provides a dependence-based prioritization lens from the upper-tail ranking that complements, and does not replace, established clinical practice and clinician judgment. Instead, we (i) reference established screening/management practices commonly used in diabetes and cardiometabolic care to aid interpretation, and (ii) highlight study-informed implications that arise specifically from our upper-tail dependence ranking, which is intended to prioritize attention toward the highest-risk strata rather than to propose new interventions.

\paragraph{CDC (Top 10 by Gumbel-$\lambda_U$: GenHlth, HighBP, DiffWalk, HighChol, BMI, HeartDiseaseorAttack, Age, PhysHlth, Stroke, CholCheck).}

\begin{enumerate}
\item \textbf{GenHlth} (self-reported general health). Poor global health concentrates in the diabetic upper tail. Implication: Use brief self-rated health screens to triage patients into accelerated diabetes risk assessment and counseling.
\item \textbf{HighBP} (hypertension). Upper-tail co-occurrence reflects cardiometabolic clustering. Implication: Integrate BP control with diabetes prevention programs and treat thresholds aggressively in high-risk clinics.
\item \textbf{DiffWalk} (difficulty walking/climbing). Functional limitation aligns with severe metabolic disease. Implication: Pair screening with referrals to physical therapy/fall-prevention and structured activity programs.
\item \textbf{HighChol }(hypercholesterolemia). Atherogenic profiles spike with diabetes risk. Implication: Integrate lipid management (statins, diet) into diabetes prevention workflows.
\item \textbf{BMI}. Extreme adiposity is strongly tail-linked to diabetes. Implication: Prioritize intensive lifestyle/weight-management services for patients above BMI cut-points.
\item \textbf{HeartDiseaseorAttack} (CHD/MI history). Macrovascular disease clusters with diabetic risk. Implication: Coordinated cardiology-endocrinology pathways and tight control of BP, lipids, and glucose.
\item \textbf{Age}. Risk concentrates in older strata. Implication: Age-adjusted screening cadence (e.g., lower testing thresholds after 45--50 years).
\item \textbf{PhysHlth}. Days of poor physical health align with high risk. Implication: Flag for care coordination and rehab referrals.
\item \textbf{Stroke}. Vascular disease clusters with diabetes. Implication: Prioritize cardio-metabolic management.
\item \textbf{CholCheck}. Screening/healthcare engagement signal. Implication: Leverage for targeted outreach.

\end{enumerate}
These signals help guide a two-step process in population health. First, brief intake items like self-rated health, blood pressure, BMI, and mobility are used to identify patients at higher risk. Next, those flagged are directed to bundled interventions that address nutrition, blood pressure and lipid control, and mobility support. Since $\lambda_U$ highlights the most extreme cases, these steps focus on patients who are most likely to face serious health issues soon.

\paragraph{PIMA (Top 8 by Gumbel--$\lambda_U$: Glucose, BMI, Age, Insulin, SkinThickness, Pregnancies, BloodPressure, DiabetesPedigreeFunction).}
\begin{enumerate}
\item \textbf{Glucose}. The dominant tail feature; high OGTT glucose co-occurs with diabetes by design and in practice. Implication: Prioritize confirmatory testing and rapid care escalation when screening glucose is extreme.
\item \textbf{BMI}. Strong adiposity signal in the upper tail. Implication: Implement intensive weight-management and lifestyle programs for patients with high BMI.
\item \textbf{Age}. Risk concentrates with advancing age. Implication: Tighten screening intervals in older adults.
\item \textbf{Insulin}. Elevated 2-hour insulin may reflect insulin resistance. Implication: Pair education on diet/activity with evaluation for metabolic syndrome.
\item \textbf{SkinThickness}. A proxy for subcutaneous fat. Implication: Reinforce anthropometric assessment when lab testing is unavailable; use as an adjunct risk flag.
\item \textbf{Pregnancies}. Parity is associated with later diabetes risk (e.g., via gestational diabetes history). Implication: Capture obstetric history and screen earlier/more often in women with multiple pregnancies or prior GDM.
\item \textbf{BloodPressure}. Hypertension co-clusters with diabetes. Implication: Integrate BP control with diabetes prevention/management visits.
\item \textbf{DiabetesPedigreeFunction}. Familial risk concentrates in the tail. Implication: Adopt earlier screening for individuals with strong family history, even at moderate BMI.
\end{enumerate}
The PIMA signals match what clinicians usually expect: glucose, BMI, and age are the most important factors. Our ranking method highlights that patients with the highest or lowest values should be prioritized for further testing and targeted risk-reduction support.

\paragraph{Summary.}
In both datasets, Gumbel-$\lambda_U$ identifies a clear group of upper-tail risk factors: global health and cardiometabolic burden in the CDC data, and glucose, adiposity, and age in the PIMA data. These findings suggest that public health efforts should focus on targeted screening and bundled interventions for people in the highest percentiles, where diabetes is most likely to co-occur and prevention can have the greatest impact.


\section*{Conclusion and Future Work}
\label{sec:conclusion}

In this study, we developed a new supervised filter method for feature selection using the Gumbel copula's upper-tail dependence coefficient, $\lambda_U$ interpreted as a rank-based tail concordance score. By focusing on how extreme values in predictors and the target label occur together, our method provides a fresh way to find features that matter most in high-risk situations. Our thorough evaluation, supports that this approach is statistically well-motivated, effective in our experiments, and computationally efficient.

We tested our method on two different diabetes datasets and found strong support for its usefulness. On the large CDC public health dataset, our Gumbel-$\lambda_U$ selector was competitive and was the fastest method, running much quicker than other approaches. It also reduced the number of features by approximately 52\% (from 21 to 10), and on this parsimonious set it achieved statistically higher performance than standard filters such as Mutual Information and mRMR, while being statistically indistinguishable from the strong ReliefF baseline. On the PIMA dataset (8 predictors), our ranking achieved the numerically highest ROC-AUC while remaining statistically indistinguishable from strong baselines, functioning as a ranking-only clinical sanity check. Although it was not the fastest on this smaller dataset, its speed was still close to other standard filters, showing that its strong performance did not come at a high computational cost. In both cases, our method consistently highlighted predictors that are clinically plausible and consistent with prior knowledge, showing it can identify meaningful and understandable signals.

This work suggests several promising directions for future research, including new methods and applications.

\paragraph{Limitation.}
The proposed screening procedure scores each predictor individually (purely marginally) via its upper-tail dependence with the outcome and therefore does not explicitly capture interaction effects among predictors. As a result, a feature that is weak marginally (or marginally uncorrelated) but informative through joint effects may be under-ranked or missed. This is a known limitation of marginal screening and is explicitly discussed in the SIS literature \cite{FanLv2008}. Group-based screening extensions have been proposed to better capture joint effects; for example, DC-SIS can be directly employed to screen grouped predictors (and multivariate responses) \cite{LiZhongZhu2012}. In future work, we plan to extend our framework to interaction-aware dependence screens (e.g., groupwise tail-dependence screening for small predictor subsets and conditional tail-dependence criteria) while retaining the interpretability and computational efficiency of the current filter.

\paragraph{Methodological Extensions.}

This work suggests many ways to extend the methodology. Since we are focused on upper-tail co-occurrence, we chose the Gumbel family with nonzero $\lambda_U$ based on its properties, instead of selecting a copula based on model fit. One next step is to look beyond the Gumbel family and compare selectors from other copulas with upper-tail dependence, such as the Joe, Student's t, or BB1 families. It may also be useful to explore new generators like the new A1 and A2 copulas \cite{Aich2025Generators, Aich2025IGNIS}, which are designed for very strong dependence. The dependence criterion could be broadened to use the lower-tail coefficient ($\lambda_L$) to find protective factors, or a combined metric for a more balanced view of risk. To make the method more rigorous, it could be combined with stability selection to control false discoveries in high-dimensional data, analyzed for finite-sample complexity, and improved with automatic feature selection using nested cross-validation or stability-based rules. Future work could also focus on making the approach more robust to covariate shift and complex missing data patterns, such as missing at random (MAR) or missing not at random (MNAR), by using copula-aware imputation. The framework could be adapted for multi-class, ordinal, and time-to-event survival data, not just binary outcomes. Finally, performance might be improved by creating hybrid and ensemble selectors that combine $\lambda_U$ scores with embedded methods like L1EN, or by using decision-analytic thresholding based on clinical utility and net benefit instead of just maximizing F1.

\paragraph{New Application Domains in Bioinformatics and Clinical Science.}

The idea of finding features that are extreme alongside high-risk outcomes can be used in many areas of biomedical research, especially where there is a lot of data. For example, in genomics and transcriptomics, this approach can help identify genes that are highly active in aggressive cancers. In proteomics and metabolomics, it can find molecules that are most abundant in severe disease or in response to treatment. This method can also improve clinical trial analysis by spotting unusual lab values that predict which patients might benefit most or be at higher risk, making it easier to focus on the right groups. In neuroimaging, it provides a new way to look at fMRI or PET scans by highlighting brain regions with unusually high activity linked to certain neurological disorders, adding to what we learn from average-based analyses. All of these uses aim to build more accurate or refined personalized risk models, helping to identify patients who are most at risk so they can get early and targeted care.

\paragraph{Practical Deployment and Ethics.}

To make this method truly effective in real-world settings, future research should tackle the challenges of practical deployment. Because important decisions often depend on spotting rare but serious risks, our model needs to be reliable when predicting extreme cases, not just accurate on average. This means we should create calibration methods that pay special attention to these rare events, so the model's confidence is justified for the highest-risk patients. We also need to carry out thorough fairness checks to make sure focusing on these extremes does not unfairly affect vulnerable groups. Building trust in the model depends on transparency, which we can support by making the selector a straightforward, repeatable preprocessing step and by recording all key details, such as random seeds, feature rankings, and confidence intervals. This approach helps ensure the process is both auditable and reproducible.

In summary, using copula-based upper-tail dependence offers an efficient and interpretable way to select features. By prioritizing predictors associated with outcomes in the upper tail and performing as well as or better than strong baselines on both large and small datasets in our experiments, this method can be a useful tool for machine learning applications in health data. The suggested extensions above outline ways to strengthen the theory, handle real-world data challenges, and evaluate the approach on additional biomedical settings.

\section*{Data availability}
The CDC Diabetes Health Indicators dataset has been accessed from the UCI Machine Learning Repository using the link \url{https://doi.org/10.24432/C53919} \cite{CDC}. The PIMA Indians Diabetes dataset originates from NIDDK and Smith et al. \cite{Smith1988}; a curated copy used for this work was accessed via Kaggle at \url{https://www.kaggle.com/datasets/uciml/pima-indians-diabetes-database}.

\section*{Code availability}
All code to reproduce the analyses is available at \url{https://github.com/agnivibes/copula-feature-selection}.

\section*{Author contributions}

A.A.: Conceived and designed the study, developed the methodology, ran the experiments, analyzed the results, and wrote the manuscript. M.M.M. and S.H.: Supported the literature review and assisted in manuscript preparation. A.M.: Interpreted the results in the public health context.

\section*{Funding}
This study was funded by the United States National Institutes of Health (NIH) under grant P20GM103434 to the West Virginia IDeA Network of Biomedical Research Excellence.

\end{document}